\documentclass[letterpaper,twocolumn,10pt]{article}
\usepackage{zhanggroup}

\usepackage{amsmath}
\usepackage{graphicx}
\usepackage{float}
\usepackage{placeins}
\usepackage{xspace}
\usepackage{tcolorbox}
\usepackage{xurl}
\usepackage{tabularx}
\newcommand{\customTableFont}{\fontsize{8pt}{8pt}\selectfont}

\newcommand{\refappendix}[1]{\hyperref[#1]{Appendix~\ref*{#1}}}
\newcommand{\mypara}[1]{\noindent\textbf{{{#1.\xspace}}}}
\newcommand{\textt}[1]{{\def\UrlFont{\ttfamily}\nolinkurl{#1}}}

\begin{document}

\date{}

\title{\bf GEO-Flag: Detecting and Measuring GEO-Optimized Web Content}

\author{
Junjie Chu\textsuperscript{1}\ \ \
Ye Leng\textsuperscript{1}\ \ \
Mingjie Li\textsuperscript{1}\ \ \
Yun Shen\textsuperscript{2}\ \ \
Xinyue Shen\textsuperscript{1,3}\ \ \
Yang Zhang\textsuperscript{1}\textsuperscript{$\clubsuit$}\ \ \
\\
\\
\textsuperscript{1}\textit{CISPA Helmholtz Center for Information Security} \ \ \
\textsuperscript{2}\textit{HPE} \ \ \
\textsuperscript{3}\textit{University of Waterloo} \ \ \
}

\maketitle
\def\thefootnote{$\clubsuit$}\footnotetext{Corresponding author.}\def\thefootnote{\arabic{footnote}}

\begin{abstract}

GEO methods apply interventions to web content to increase its likelihood of being selected and cited by generative search engines.
This can give strategically optimized pages visibility disproportionate to their authority or relevance and even make weak or false information appear well supported.
Unlike conventional search, generative search synthesizes information into direct answers rather than presenting competing sources, which can further amplify these risks, as assessing source provenance and authority requires additional user interaction.
Despite these concerns, systematic methods for detecting GEO-optimized webpages remain underexplored.
We introduce \texttt{GEOFlagBench}, a benchmark of 3,200 web content instances spanning 400 queries, four domains, and eight GEO optimizer families, and use it to systematically evaluate existing GEO detection methods.
Although the strongest baseline achieves an aggregate F1 of 0.880, method-level and authorship-conditioned evaluations reveal substantial weaknesses and potential reliance on authorship-related shortcuts.
We therefore propose \emph{Intervention-Paired Training} (IPT), which supervises detector responses to GEO interventions and non-GEO AI polishing; on ModernBERT, IPT improves F1 from 0.862 to 0.944 and worst-group accuracy from 0.725 to 0.883.
Beyond page-level detection, we develop a GEO-gated Agent system for auditing the URL source tier and verifiability of citation URLs used by detected GEO pages.
Finally, we deploy the complete pipeline on released Google Search and Gemini-grounded retrieval results for 1,000 real-user queries.
Across 10,095 available pages, we estimate an overall GEO prevalence of 8.90\%, reaching 16.36\% among pages modified in 2026, while 69.34\% of citation occurrences on detected GEO pages receive LOW verifiability labels.
Our results establish a foundation for systematically detecting, auditing, and measuring GEO in real-world search ecosystems.

\end{abstract}

\section{Introduction}

Generative search engines, such as Perplexity and Google Gemini-Grounded Search, are emerging as a new way to access information on the web.
By retrieving multiple pages and synthesizing them into a single answer with inline citations, they make information access more convenient and efficient.
Their growing adoption has also given rise to \emph{Generative Engine Optimization} (GEO), which applies interventions to web content to increase its likelihood of being selected and cited by generative search engines~\cite{AMRKND24}.
GEO has already developed into a growing commercial optimization practice.

\begin{figure}[!t]
\centering
\includegraphics[width=1.0\columnwidth]{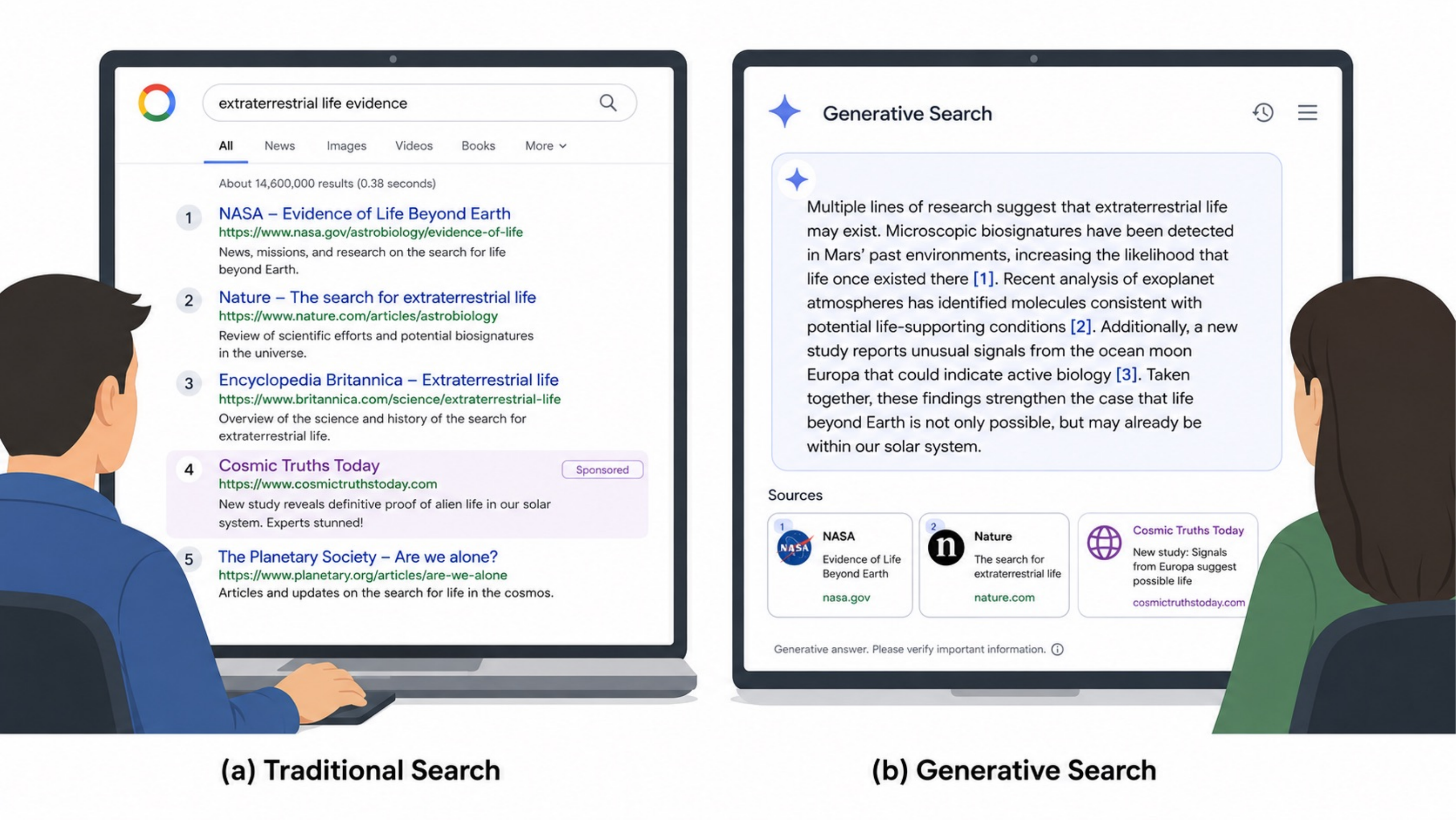}
\caption{Illustration of how generative search can amplify risks from GEO-optimized or strategically constructed content.
Traditional search exposes multiple competing sources and their provenance, whereas generative search synthesizes information into a single answer with less visible source context.}
\label{figure:trad_ge_compare}
\end{figure}

However, GEO also introduces new risks.
Highly optimized pages may receive visibility disproportionate to their actual authority or relevance, potentially displacing more reliable sources.
Worse, GEO techniques can be misused to make weak or false information appear well supported.
For example, an operator can publish a false claim on one website and then cite that page from another~\cite{fan2026geo,cna2026geo}; the page URL is valid, but the underlying source was strategically created to support the claim itself.
Generative search can even \textbf{amplify} these risks.
Unlike conventional search, it presents users with a synthesized answer rather than a list of competing sources, making the provenance and authority of the underlying information less immediately visible (see~\autoref{figure:trad_ge_compare}).
As a result, claims from strategically constructed sources may spread more easily, as GEO techniques and generative search can make users less willing to inspect and verify the underlying sources.

Despite these risks, there is still a lack of systematic tools for identifying GEO-optimized content.
Existing work primarily studies how to improve visibility in generative engines~\cite{AMRKND24,PBGS24,NDT25,WZKX25,PGGOY25}, while considerably less attention has been paid to the inverse problem: determining whether a webpage has been deliberately optimized for generative search.

\mypara{Measuring Current Detection Methods}
To fill this gap, we start by building a new benchmark dataset and measuring current available detection methods.
Concretely, we first propose \texttt{GEOFlagBench}, a benchmark for detecting GEO-optimized web content across diverse domains and optimization strategies.
It contains 3{,}200 web content instances derived from 400 queries across four domains and spans eight GEO optimizer families.
We then utilize \texttt{GEOFlagBench} to evaluate various GEO detectors, including fine-tuned classifiers, zero-shot LLMs, feature-based models, and repurposed AI-text detectors.
Although the strongest baseline reaches an aggregate F1 of 0.880, further evaluation on methods and authorship reveals substantial weaknesses.
For example, Word TF-IDF has a worst-group accuracy of only 0.375 and a false-positive-rate gap of 0.558 between AI- and human-authored non-GEO instances.
These results suggest that high aggregate performance can conceal reliance on original-authorship or general LLM-writing cues instead of GEO-specific interventions.

\mypara{Improved GEO Detection}
To address these limitations, we propose \textit{Intervention-Paired Training (IPT)}, which directly supervises how a detector should respond to documented content transformations.
IPT uses positive pairs to require a higher GEO score after a GEO intervention and zero pairs to preserve the score after non-GEO AI polishing.
This design encourages the detector to capture optimization-specific changes while reducing its reliance on potential shortcuts, such as static authorship and generic LLM-editing signals.
On ModernBERT, IPT improves F1 from 0.862 to 0.944 and worst-group accuracy from 0.725 to 0.883, while reducing the authorship-conditioned false-positive-rate gap from 0.263 to 0.108.
IPT also improves Qwen's F1 from 0.833 to 0.883 and its worst-group accuracy from 0.325 to 0.775.

Beyond detecting whether the content of a webpage is GEO-optimized, we further develop a GEO-gated Agent system for assessing the verifiability of citation URLs used by detected GEO webpages.
The system first applies the GEO detector to identify GEO-optimized pages and deterministically extracts the citation URLs.
Then, for each citation URL, a constrained Agent assesses its URL source tier, collects its accessibility level, and derives citation URL verifiability via deterministic rules.
We evaluate the system on an independent benchmark containing 562 web content instances, organized into 281 non-GEO/GEO pairs.
Among correctly detected GEO instances, the strongest Agent, GLM 5.2, achieves 84.75\% accuracy for URL source tier and 83.00\% for citation URL verifiability.

\mypara{Empirical GEO Prevalence Estimation}
We then deploy the complete GEO detection and citation URL audit pipeline to estimate GEO prevalence in real-world search results.
We analyze pages associated with 1{,}000 real-user ORCAS queries~\cite{CCMYB20,GLCSBC26} in released conventional Google Search and Gemini-grounded retrieval results.
After multi-stage page collection and recovery, the audit contains 10{,}095 usable pages from 13{,}985 unique URLs.
The detector flags 898 pages as GEO, corresponding to an estimated prevalence of 8.90\%, with 8.14\% for conventional Google search and 9.09\% for Gemini-grounded search.
Among pages with a parseable declared \texttt{dateModified} value, the estimated GEO prevalence in the unique union increases from 7.02\% in 2024 to 12.80\% in 2025 and 16.36\% in 2026.
Among 6{,}663 citation occurrences extracted from the detected GEO pages, 69.34\% receive LOW verifiability labels, meaning that the cited sources offer limited editorial accountability or are difficult for readers to independently inspect.
The LOW share reaches 74.15\% for Gemini, compared with 45.88\% for Google Search, indicating substantial differences in the citation URL composition of GEO pages exposed through the two channels.\footnote{
For live webpages, we cannot determine with certainty whether a page has actually undergone GEO, and the detection pipeline is not error-free.
We therefore report these findings as empirical estimates rather than definitive measurements.
}

\mypara{Contributions}
Our contributions are summarized as follows:
\begin{itemize}
\item We introduce \texttt{GEOFlagBench} and use it to systematically evaluate current GEO detection methods, revealing substantial method-level weaknesses and potential reliance on authorship-related shortcuts despite strong aggregate performance.
\item We propose \emph{Intervention-Paired Training} (IPT), which improves GEO detection while reducing reliance on authorship and generic AI-writing cues.
\item We develop a GEO-gated Agent system for auditing the URL source tier and verifiability of citation URLs used by detected GEO pages.
\item We provide an empirical estimate of GEO prevalence in released Google Search and Gemini-grounded retrieval results, finding an overall estimated prevalence of 8.90\% and a clear increase among more recently modified pages, reaching 16.36\% among pages modified in 2026.
\end{itemize}

\section{GEO Flagging with Existing Detectors}
\label{section:current_methods}

\begin{figure}[!t]
\centering
\includegraphics[width=.96\columnwidth]{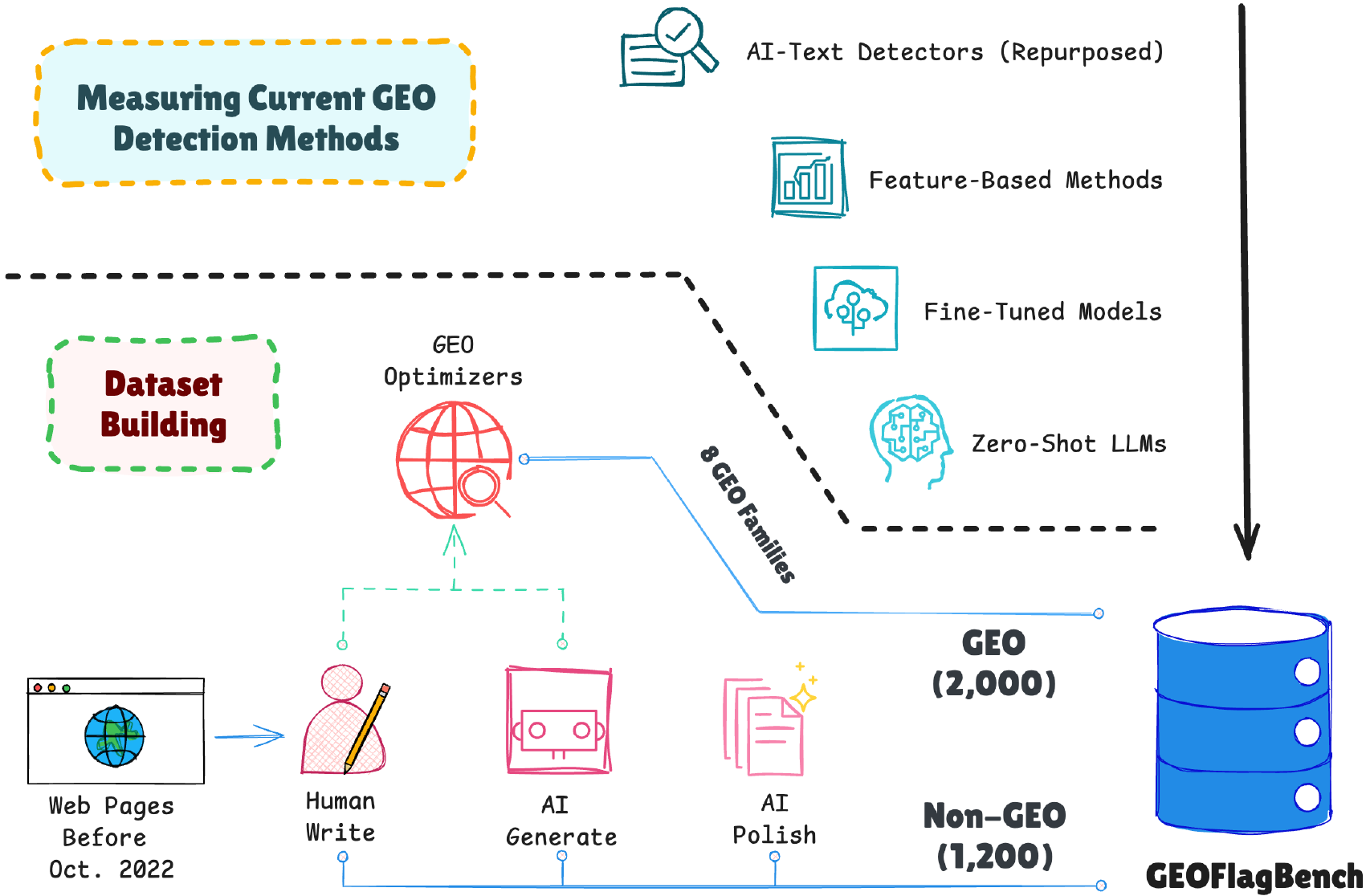}
\caption{Overview of dataset construction and measurement of the existing GEO detection methods.}
\label{figure:geoflagbench}
\end{figure}

We begin by measuring how well existing approaches distinguish GEO-optimized page content from non-GEO content.
The entire pipeline of benchmark construction and current method measurement is outlined in~\autoref{figure:geoflagbench}.
This serves two goals.
First, we measure how current approaches perform in a controlled, query-disjoint evaluation.
Second, we examine whether detectors perform consistently across different GEO methods and on instances derived from human- and AI-authored content.
The insights subsequently motivate the Intervention-Paired Training method later introduced in~\autoref{section:ipt}.

\subsection{Task Definition and Evaluation Scope}
\label{section:task_formulation}

We define \emph{GEO detection} as the binary task of determining whether a web content instance has undergone a GEO intervention.
Given a web content instance $x$, a detector predicts $\hat{y}\in\{0,1\}$, where $\hat{y}=1$ means that the instance was GEO-optimized to improve its visibility, ranking, or selection by a generative search system, and $\hat{y}=0$ means that it was not.
It is important to bear in mind that our goal is to detect the presence of a GEO-specific intervention, not authorship or writing style.
A GEO web content instance may be derived from either human-written or AI-generated source content, while a non-GEO instance may likewise be human-written, AI-polished, or AI-generated.
A useful detector must distinguish GEO-specific signals from general effects of AI generation, editing, query, and source provenance.

In this paper, we consider an instance-only input setting.
The detector sees the resulting content instance but not its original version, construction prompt, or provenance metadata.

\subsection{\texttt{GEOFlagBench}}
\label{section:geoflagbench}

\mypara{Overview}
\texttt{GEOFlagBench} contains 3{,}200 web content instances constructed from 400 queries across Health, Finance, Technology, and Travel.
For each query, we construct three non-GEO controls: a human-written source web content instance, an AI-polished version of that instance, and an AI-generated web content instance derived from a structured summary of the source instance.
This process yields 1{,}200 non-GEO controls.
We then select human-written and AI-generated controls as seeds for eight GEO optimizer families.
Each application of an optimizer produces a separate GEO web content instance, yielding 2{,}000 GEO instances in total.
Note that AI-polished controls are never used as GEO seeds.
The full details are reported in~\autoref{table:geoflagbench_composition}.

\mypara{Domain Selection and Query Collection}
We construct the benchmark across four domains: Health, Finance, Technology, and Travel.
These domains represent heterogeneous information environments in which GEO effectiveness may vary~\cite{AMRKND24}.
Health and Finance cover consequential settings in which inaccurate information may affect users' health or financial well-being, while Technology captures rapidly evolving technical content and Travel captures practical consumer information~\cite{SGSZ25,google_quality_guidelines}.
We include 100 queries per domain to maintain a balanced domain distribution and prevent domain frequency from becoming a shortcut for classification.

We focus on general informational queries that seek factual or explanatory content without targeting a specific website, platform, or other destination.
Destination-specific queries are excluded because they strongly constrain the expected source and page format, whereas general informational queries allow the same query to be addressed by content from different publishers.
We first collect candidate queries from the GEO-Bench test split using its released annotations~\cite{AMRKND24}.
We use the test split for two reasons.
First, its fine-grained annotations allow us to exclude sensitive queries and systematically map the remaining candidates to our four domains.
Second, the GEO-Bench training split has been used to develop or train downstream GEO optimization methods~\cite{AMRKND24,WZKX25}.
Using held-out test queries therefore reduces direct query-level overlap with the data used to develop the optimization methods evaluated in our benchmark.
We then apply Sonnet 4.6 to screen the mapped queries for domain mismatch and semantic duplication, leaving 164 GEO-Bench queries.

To obtain 100 queries per domain, we use Sonnet 4.6 to generate 236 additional informational queries under fixed domain and quality constraints.
Finally, two authors independently review all 400 queries for domain correctness and semantic duplication.
Both confirm that every query belongs to its assigned domain and that no semantic duplicates remain.
Further details on the domain definitions, tag-to-domain mapping, filtering criteria, and query generation process are provided in \refappendix{section:domain_query_details}.

\mypara{Non-GEO Data Construction}
For each of the 400 queries above, we construct three non-GEO groups: human-written, AI-polished, and AI-generated.
\begin{itemize}
    \item \textbf{Human-Written.}
    We retrieve one webpage per query from English Wikipedia or relatively reliable publishers returned by the Brave Search API.
    To reduce the possibility of prior GEO intervention, we restrict all sources to content available before October 1, 2022, which predates both the formal introduction of GEO~\cite{AMRKND24} and the public release of ChatGPT.
    We further verify modification metadata and retain only pages with no recorded updates after this cutoff.
    This yields 400 human-written non-GEO seed instances.
    \item \textbf{AI-Polished.}
    For each human-written seed, we create a version in which an LLM improves grammar, spelling, punctuation, and phrasing while preserving the original semantics and content.
    We distribute the instances across eight different LLMs to reduce model-specific artifacts from becoming classification shortcuts.
    The prompts contain no GEO-related instructions, allowing this group to isolate the effect of AI-assisted editing.
    \item \textbf{AI-Generated.}
    For each query, we first extract a structured summary from the corresponding human-written instance and then ask an LLM to write a new article from that summary while preserving its main factual content and targeting a similar length.
    The 400 queries are again distributed across the same eight LLMs.
    Because the generation prompts contain no GEO objectives, this group serves as a control for fully AI-generated content without deliberate GEO optimization.
\end{itemize}
The detailed construction process for each non-GEO group is described in~\refappendix{section:non-geo_details}.

\mypara{GEO Data Construction}
We construct 2{,}000 GEO web content instances by applying explicit optimization interventions to human-written or AI-generated non-GEO seeds.
The benchmark includes eight optimizer families that cover different GEO mechanisms and levels of intervention: GEO Strategy Pool~\cite{AMRKND24}, AutoGEO~\cite{WZKX25}, PMA~\cite{NDT25}, RAID G-SEO~\cite{CWBCLH25}, Meta-Optimization~\cite{BFKPW25}, Stealthy GEO Strategy Pool, Stealthy AutoGEO, and Human GEO.
The first four families cover strategy-based rewriting, learned optimization preferences, preference manipulation, and intent-aware optimization, respectively.
Meta-Optimization adds iterative self-evaluation and revision.
We further construct stealthy variants of GEO Strategy Pool and AutoGEO that retain the same GEO objectives while reducing obvious optimization patterns in the resulting text.
Finally, we conduct Human GEO, which combines various GEO strategies and is applied by human annotators.
Where applicable, we use multiple executor LLMs to reduce the risk that a detector learns model-specific generation artifacts instead of features associated with GEO interventions.
The complete construction process for each GEO family is described in~\autoref{table:geo-families} of~\refappendix{section:geo_details}.

\subsection{Experiment Settings}
\label{section:geoflagbench_setting}

\mypara{Train and Test Split}
To prevent semantic leakage (e.g., web content instances derived from the same query appearing in both training and testing) and evaluate generalization to unseen queries, we use a query-level grouped split, assigning all instances associated with the same query to the same partition.
Concretely, we apply stratified sampling by domain, assigning 70 queries per domain to training and 30 queries per domain to testing.
The data associated with these queries form the training and test sets, containing 2{,}238 and 962 instances, respectively.

\mypara{Metrics}
We report accuracy and F1 on the fixed test set.
For binary tasks, F1 treats GEO as the positive class.
For multiclass tasks, we report macro F1, computed as the unweighted mean of the per-class F1 scores.
Probabilistic classifiers use a decision threshold of 0.5.
The proprietary APIs use the label mappings stated below.

\subsection{Existing Detectors}
\label{section:current_method_settings}

\mypara{Fine-Tuned Models}
We fine-tune ModernBERT-base~\cite{WCCWHTGBLAAHP25} as a binary GEO classifier with context lengths of 1{,}024 and 8{,}192 tokens.
These models measure how well a classifier trained on webpage text can predict the benchmark labels.
We also fine-tune Qwen3-0.6B~\cite{qwen3} using a structured output that contains an intermediate analysis followed by a binary decision (\textt{Answer: Yes/No}).
Only the final decision is used for evaluation.
We fine-tune all ModernBERT models for 5 epochs, and Qwen models for 10 epochs with standard supervised fine-tuning (SFT)~\cite{DCLT19,LOGDJCLLZS19}.

\mypara{Feature Engineering and Classical Classifiers}
For lexical features, we train logistic regression (LR) on character TF-IDF features with 2--4 character grams and word TF-IDF features with unigrams and bigrams, retaining at most 20{,}000 features for each representation.
We additionally train logistic regression on Ghostbuster features~\cite{VFTK24} and on GPT-2 perplexity-derived features~\cite{RWCLAS19,AMRKND24}.
All logistic regression models use an inverse regularization strength of $C=1$ and a maximum of 1{,}000 optimization iterations.
To capture webpage structure, we train XGBoost~\cite{CG16} on ten structural features, including content length and the frequencies of headings, lists, citations, numbers, and sections.
The XGBoost classifier uses 200 trees, a maximum depth of 6, and a learning rate of 0.1.

\mypara{Proprietary AI-Text Detector APIs}
We evaluate two commercial AI-text detection services, Pangram~\cite{pangram_detector} and GPTZero~\cite{gptzero_detector}, using their standard interfaces.
GPTZero returns three instance-level classes: AI, Mixed, and Human.
We thus evaluate two mappings to the GEO label: GPTZero treats only AI as GEO, while GPTZero (mixed) treats both AI and Mixed as GEO.

\mypara{Zero-Shot LLMs}
We evaluate Qwen3-0.6B~\cite{qwen3}, Gemini 3.5 Flash~\cite{gemini35flash}, and Claude Haiku 4.5~\cite{Claude-Haiku-4.5} as zero-shot GEO detectors.
Each model receives the same task definition and web content instance, without demonstrations, and returns a single GEO or non-GEO decision.

\mypara{AI-Text Detectors Repurposed for GEO}
Finally, we evaluate several existing AI-text detection signals after adapting their outputs to the GEO task.
These include the off-the-shelf Ghostbuster score~\cite{VFTK24}, the Fast-DetectGPT criterion~\cite{BZTYZ24}, and the machine-generated probability ($p_{\text{machine}}$) from the RoBERTa OpenAI Detector~\cite{openai_detector}.
We also include maximum keyword term frequency as a simple lexical GEO signal.
Because these methods were not designed to detect GEO, their native scores do not directly correspond to the GEO label.
For each score, we therefore fit a one-dimensional logistic regression using only the training split.
This calibration changes only the mapping from the native score to the GEO label and leaves the underlying detector or representation unchanged.

\subsection{Evaluation}
\label{section:current_method_results}

\begin{table}[!t]
\centering
\caption{Overall GEO flagging results on the fixed test set of 360 non-GEO and 602 GEO instances.
The last block calibrates each native score with one-dimensional logistic regression on the training split.
}
\label{table:baselines}
\setlength{\tabcolsep}{3pt}
\customTableFont
\begin{tabular}{lcc}
\toprule
\textbf{Method} & \textbf{Acc.} & \textbf{F1} \\
\midrule
\multicolumn{3}{l}{\textbf{\textit{Fine-tuned Models}}} \\
ModernBERT (1K)  & 0.842 & 0.864 \\
ModernBERT (8K)     & 0.839 & 0.862 \\
Qwen3-0.6B (SFT)               & 0.778 & 0.833 \\
\midrule
\multicolumn{3}{l}{\textbf{\textit{Feature engineering + classical classifiers}}} \\
TF-IDF char (2-4) + LR              & 0.848 & 0.878 \\
TF-IDF word (1-2) + LR              & 0.846 & 0.880 \\
Ghostbuster features, recalibrated  & 0.796 & 0.835 \\
GPT-2 perplexity + LR               & 0.710 & 0.781 \\
Structural + XGBoost                & 0.752 & 0.801 \\

\midrule
\multicolumn{3}{l}{\textbf{\textit{Proprietary AI-text detector API}}} \\
Pangram              & 0.838 & 0.876 \\
GPTZero (mixed)      & 0.799 & 0.830 \\
GPTZero               & 0.768 & 0.782 \\
\midrule
\multicolumn{3}{l}{\textbf{\textit{Zero-shot LLMs}}} \\
Qwen3-0.6B             & 0.627 & 0.770 \\
Gemini-3.5-flash       & 0.759 & 0.776 \\
Haiku-4.5              & 0.747 & 0.771 \\
\midrule
\multicolumn{3}{l}{\textbf{\textit{AI-text detectors repurposed for GEO (train-calibrated)}}} \\
Ghostbuster (off-shelf score) + LR          & 0.626 & 0.770 \\
Fast-DetectGPT (criterion) + LR                  & 0.631 & 0.767 \\
RoBERTa OpenAI ($p_{\text{machine}}$) + LR    & 0.626 & 0.770 \\
Keyword max TF + LR                         & 0.626 & 0.770 \\
\bottomrule
\end{tabular}
\end{table}

\begin{table*}[!ht]
\centering
\caption{Method-level recall and authorship-conditioned diagnostics.
Arrows indicate the preferred direction.}
\label{table:shortcut_metrics}
\setlength{\tabcolsep}{3pt}
\customTableFont
\begin{tabular}{lcccccc}
\toprule
\textbf{Method}
& \multicolumn{3}{c}{\textbf{Method-Level GEO Recall}}
& \multicolumn{3}{c}{\textbf{Authorship-Conditioned Diagnostics}} \\
\cmidrule(lr){2-4}\cmidrule(lr){5-7}
& \textbf{PMA$\uparrow$} & \textbf{AutoGEO-Light$\uparrow$}
& \textbf{Human$\uparrow$}
& \textbf{Worst-Group Acc.$\uparrow$}
& \textbf{$\Delta$FPR$\downarrow$} & \textbf{$\Delta$TPR$\downarrow$} \\
\midrule
\multicolumn{7}{l}{\textbf{\textit{Fine-tuned models}}} \\
ModernBERT-base (1K)  & 0.566 & 0.423 & 0.426 & 0.725 & 0.271 & 0.121 \\
ModernBERT-base (8K)     & 0.513 & 0.346 & 0.444 & 0.725 & 0.263 & 0.117 \\
Qwen3-0.6B (SFT)               & 0.934 & 0.654 & 0.463 & 0.325 & 0.413 & 0.105 \\
\midrule
\multicolumn{7}{l}{\textbf{\textit{Feature engineering + classical classifiers}}} \\
TF-IDF char (2-4) + LR              & 0.737 & 0.308 & 0.778 & 0.575 & 0.346 & 0.077 \\
TF-IDF word (1-2) + LR              & 0.908 & 0.423 & 0.704 & 0.375 & 0.558 & 0.111 \\
Ghostbuster features, recalibrated  & 0.539 & 0.192 & 0.759 & 0.600 & 0.221 & 0.047 \\
Structural + XGBoost                & 0.421 & 0.538 & 0.759 & 0.550 & 0.188 & 0.033 \\
GPT-2 perplexity + LR               & 0.605 & 0.538 & 0.685 & 0.367 & 0.221 & 0.084 \\
\midrule
\multicolumn{7}{l}{\textbf{\textit{Proprietary AI-text detector API}}} \\
Pangram              & 1.000 & 0.769 & 0.500 & 0.192 & 0.767 & 0.099 \\
GPTZero (mixed)      & 1.000 & 0.462 & 0.389 & 0.508 & 0.475 & 0.185 \\
GPTZero               & 0.961 & 0.192 & 0.074 & 0.583 & 0.154 & 0.172 \\
\midrule
\multicolumn{7}{l}{\textbf{\textit{Zero-shot LLMs}}} \\
Qwen3-0.6B             & 1.000 & 1.000 & 1.000 & 0.008 & 0.000 & 0.006 \\
Gemini-3.5-flash       & 0.987 & 0.154 & 0.037 & 0.589 & 0.258 & 0.166 \\
Haiku-4.5              & 0.987 & 0.269 & 0.074 & 0.586 & 0.321 & 0.197 \\
\midrule
\multicolumn{7}{l}{\textbf{\textit{AI-text detectors repurposed for GEO (train-calibrated)}}} \\
Ghostbuster (off-shelf score) + LR          & 1.000 & 1.000 & 1.000 & 0.000 & 0.000 & 0.000 \\
Fast-DetectGPT (criterion) + LR                 & 0.961 & 0.962 & 1.000 & 0.004 & 0.179 & 0.037 \\
RoBERTa OpenAI ($p_{\text{machine}}$) + LR    & 1.000 & 1.000 & 1.000 & 0.000 & 0.004 & 0.003 \\
Keyword max TF + LR                         & 1.000 & 1.000 & 1.000 & 0.000 & 0.000 & 0.000 \\

\bottomrule
\end{tabular}
\end{table*}

\mypara{Overall Results}
As shown in~\autoref{table:baselines}, several methods perform competitively on the fixed test set.
Word TF-IDF with logistic regression achieves the highest F1 of $0.880$, while ModernBERT (8K) also obtains a high accuracy of $0.839$ with an F1 of $0.862$.
Character TF-IDF and Pangram perform similarly, reaching F1 scores of $0.878$ and $0.876$, respectively.
These results show that competitive GEO flagging is possible with substantially different approaches, ranging from fine-tuned encoders to simple lexical features and a proprietary AI-text detector.

Other approaches are considerably less effective.
The three zero-shot LLMs obtain F1 scores between $0.770$ and $0.776$, while several repurposed AI-text detector scores remain close to a positive-class majority baseline even after calibration on the training split.
This baseline predicts every instance as GEO and obtains an accuracy of $0.626$ and an F1 of $0.770$.

Overall, the aggregate results suggest that GEO interventions leave detectable signals, but performance varies substantially across detector families.

\mypara{Method-Level Results}
Aggregate metrics, however, do not show whether a high-performing detector works consistently across different GEO methods.
We therefore examine method-level recall values.
We list the results of three representative methods (PMA, AutoGEO-Light, and Human GEO) in~\autoref{table:shortcut_metrics}.
These subsets represent substantially different forms of optimization: PMA introduces explicit ranking-oriented interventions, AutoGEO-Light applies relatively sparse interventions, and Human GEO is considered to have fewer common AI-writing cues.
Despite their strong aggregate performance, several detectors show pronounced differences across these methods.
The two TF-IDF classifiers exhibit a similar imbalance.
Although they achieve the two highest aggregate F1 scores in~\autoref{table:baselines}, their AutoGEO-Light recall is only $0.308$ and $0.423$ in~\autoref{table:shortcut_metrics}.
The contrast is even stronger for some zero-shot and proprietary methods: Gemini and Haiku both reach $0.987$ recall on PMA, but only $0.037$ and $0.074$ on Human GEO, respectively.
GPTZero likewise reaches only $0.074$ recall on Human GEO.

The three subsets also suggest different sources of difficulty.
PMA is detected reliably by many methods, possibly because its more explicit optimization introduces stronger lexical or stylistic signals.
AutoGEO-Light presents the opposite case: because it intervenes on only a limited portion of the instance, the resulting GEO signal may be diluted by the largely unchanged source text.
Its low recall therefore suggests that some detectors are sensitive to the extent of optimization rather than merely to its presence.
Human GEO raises a different concern.
It is designed to reduce stylistic cues commonly associated with AI-written text.
The sharp recall degradation on this subset therefore raises the possibility that some detectors rely partly on human--AI original authorship cues rather than signals specific to GEO itself.

\mypara{Human--AI Original Authorship as a Potential Shortcut}
The low recall on Human GEO raises the possibility that some detectors rely on cues associated with the original human--AI authorship of an instance rather than GEO itself.

To examine this possibility, we partition the test set into four groups defined by GEO status and \emph{original authorship}: human non-GEO, AI non-GEO, human GEO, and AI GEO.
Here, original authorship refers to the provenance of the underlying source content before any polishing or GEO transformation.
Accordingly, AI-polished instances are included in the human non-GEO group because they originate from human-written webpages and undergo only grammar- and fluency-level editing without any GEO strategy.
The resulting groups contain 240 human non-GEO, 120 AI non-GEO, 321 human GEO, and 281 AI GEO instances.
We then use three complementary diagnostics in~\autoref{table:shortcut_metrics} to measure how strongly detector performance varies across these original-authorship groups:
\begin{itemize}
\item \textbf{Worst-group accuracy (WGA)} is the minimum accuracy across the four groups~\cite{SKHL20}.
It measures whether strong aggregate performance is maintained for every combination of GEO status and authorship.
Higher values are better; a low value indicates that aggregate accuracy masks poor performance on at least one group.
\item \textbf{$\Delta$FPR} is the absolute difference in false-positive rate between AI and human non-GEO instances~\cite{HPS16}.
It measures whether authorship affects the tendency to incorrectly flag a non-GEO instance as GEO.
Lower values are better; a large gap means that the detector treats human- and AI-authored non-GEO instances differently despite their identical GEO label.
\item \textbf{$\Delta$TPR} is the absolute difference in true-positive rate between AI and human GEO instances~\cite{HPS16}.
It measures whether authorship affects the detector's ability to recognize GEO instances.
Lower values are better; a large gap means that GEO detection success depends substantially on whether the underlying instance is human- or AI-authored.
\end{itemize}

Several methods with strong aggregate performance exhibit substantial authorship-conditioned differences.
Word TF-IDF achieves the highest overall F1 of $0.880$, yet its worst-group accuracy is only $0.375$ and its $\Delta$FPR reaches $0.558$ in~\autoref{table:shortcut_metrics}.
Pangram shows an even larger discrepancy: despite an aggregate F1 of $0.876$, its worst-group accuracy is $0.192$ and its $\Delta$FPR is $0.767$.
Qwen3-0.6B (SFT) similarly combines an F1 of $0.833$ with a worst-group accuracy of $0.325$ and a $\Delta$FPR of $0.413$.
Thus, some of the methods that appear strongest under aggregate evaluation are substantially less stable once human and AI authorship are considered separately.

These results are consistent with human--AI authorship acting as a shortcut for GEO flagging.
A detector may partly distinguish AI-written from human-written instances rather than identifying signals introduced specifically by GEO.
Our analysis is observational and does not establish the internal decision rule of any detector, but the large authorship-conditioned gaps show that aggregate performance alone cannot rule out such shortcut reliance.

Small authorship gaps must also be interpreted with care.
Zero-shot Qwen3-0.6B and several calibrated AI-text scores predict nearly every instance as GEO, producing near-zero authorship gaps and near-perfect recall on the listed GEO subsets but worst-group accuracy close to zero.
Their apparent stability therefore results from nearly constant predictions rather than robust GEO detection.

These results reveal that \emph{existing GEO detectors may vary substantially across optimization methods, and some may rely partly on human--AI authorship cues rather than GEO-specific signals}.

\section{Toward More Reliable GEO Detection}
\label{section:ipt}

\subsection{Motivation}

Our analysis in~\autoref{section:current_method_results} reveals two limitations of existing GEO detectors.
First, some high-performing detectors generalize poorly across GEO methods, especially those with sparse interventions such as AutoGEO-Light.
Second, their error rates can differ substantially between instances derived from human- and AI-authored sources, suggesting potential reliance on original-authorship shortcuts rather than GEO-specific signals.

To directly address these two limitations, we propose \emph{Intervention-Paired Training} (IPT), which exploits the correspondence between original and transformed instances in \texttt{GEOFlagBench}.
IPT is designed to make the detector sensitive to GEO interventions while insensitive to changes that are unrelated to GEO.
For instances derived from the same source, a GEO intervention should increase the GEO score, which encourages the detector to recognize even small GEO-induced changes rather than relying on patterns specific to particular GEO methods.
In contrast, non-GEO AI polishing should leave the score approximately unchanged, discouraging the detector from treating generic LLM processing as evidence of GEO.
Because original authorship is fixed within each pair, IPT also reduces the usefulness of authorship-related cues and encourages the detector to rely on GEO-specific signals.\footnote{We additionally evaluate a Gradient Reversal Layer (GRL) control that explicitly suppresses original-authorship information in the learned representation; see~\refappendix{appendix:grl}.}

\subsection{Intervention-Paired Training}

\mypara{Positive and Zero Interventions}
We construct two types of intervention pairs.
\begin{itemize}
\item A \emph{positive pair} $(x_o,x_g)$ consists of an original instance $x_o$ and its GEO-optimized counterpart $x_g$.
Because the transformation from $x_o$ to $x_g$ applies a GEO intervention, we require the GEO score of $x_g$ to exceed that of $x_o$ by a predefined margin.
This constraint encourages the detector to capture changes introduced by GEO optimization rather than relying only on static properties of individual instances.

\item A \emph{zero pair} $(x_o,x_p)$ consists of an original instance $x_o$ and its AI-polished counterpart $x_p$.
AI polishing introduces LLM-induced lexical and stylistic changes but does not apply a GEO strategy.
We therefore constrain the GEO scores of $x_o$ and $x_p$ to remain close, explicitly teaching the detector that LLM editing alone is not evidence of GEO.
\end{itemize}

\noindent
The two pair types provide complementary supervision.
Positive pairs identify changes that should increase the GEO score, whereas zero pairs identify LLM-induced changes that should not affect it.

\mypara{IPT Objective}
Let $z_{\theta}(x)$ denote the binary GEO logit for instance $x$, and let $\mathcal{D}$ denote the labeled training instances.
The standard instance-level classification loss is
\begin{equation}
\mathcal{L}_{\mathrm{cls}}
=
\frac{1}{|\mathcal{D}|}
\sum_{(x,y)\in\mathcal{D}}
\operatorname{BCE}\!\left(
\sigma\!\left(z_{\theta}(x)\right), y
\right),
\label{equation:ipt_cls}
\end{equation}
where $\sigma(\cdot)$ is the sigmoid function and BCE is the binary cross entropy.
Let $\mathcal{P}_{+}$ denote the set of positive intervention pairs.
We require a GEO intervention to increase the GEO logit by at least margin $m$:
\begin{equation}
\mathcal{L}_{+}
=
\frac{1}{|\mathcal{P}_{+}|}
\sum_{(x_o,x_g)\in\mathcal{P}_{+}}
\max\!\left(
0,
m-z_{\theta}(x_g)+z_{\theta}(x_o)
\right).
\label{equation:ipt_positive}
\end{equation}

\noindent
Let $\mathcal{P}_{0}$ denote the set of zero intervention pairs.
For these pairs, we constrain AI polishing to preserve the GEO logit:
\begin{equation}
\mathcal{L}_{0}
=
\frac{1}{|\mathcal{P}_{0}|}
\sum_{(x_o,x_p)\in\mathcal{P}_{0}}
\left|
z_{\theta}(x_p)-z_{\theta}(x_o)
\right|.
\label{equation:ipt_zero}
\end{equation}

\noindent
The complete binary IPT objective is
\begin{equation}
\mathcal{L}_{\mathrm{IPT}}
=
\mathcal{L}_{\mathrm{cls}}
+
\alpha \mathcal{L}_{+}
+
\beta \mathcal{L}_{0}.
\label{equation:ipt_total}
\end{equation}

\noindent
IPT does not require every training instance to belong to an intervention pair.
Unpaired instances receive only the standard classification loss in~\autoref{equation:ipt_cls} and therefore reduce to ordinary supervised fine-tuning.
IPT can thus be applied as an extension of standard SFT: paired instances provide additional intervention-level supervision, while unpaired instances retain the standard instance-level objective.
Pair construction uses only training queries, and inference remains instance-based and requires no paired input.

\mypara{Generative IPT}
We further apply IPT to Qwen3-0.6B using LoRA adapters~\cite{HSWALWWC22}.
The model generates a reasoning sequence followed by a Yes/No verdict.
We define the GEO score as
\begin{equation}
z_{\theta}(x)
=
\ell_{\theta}(\mathrm{Yes}\mid x)
-
\ell_{\theta}(\mathrm{No}\mid x),
\label{equation:ipt_qwen_score}
\end{equation}
where the two logits are evaluated at the verdict position.
For notational simplicity, $z_{\theta}(x)$ denotes this verdict-position logit difference, including the preceding reasoning context.

The generative objective follows the same formulation as binary IPT.
We replace the instance-level classification loss $\mathcal{L}_{\mathrm{cls}}$ with the completion-only language-modeling loss $\mathcal{L}_{\mathrm{LM}}$ over the reasoning sequence and final verdict:
\begin{equation}
\mathcal{L}_{\mathrm{IPT}}^{\mathrm{gen}}
=
\mathcal{L}_{\mathrm{LM}}
+
\alpha \mathcal{L}_{+}
+
\beta \mathcal{L}_{0}.
\label{equation:ipt_gen_total}
\end{equation}

The positive and zero losses retain the definitions in~\autoref{equation:ipt_positive} and~\autoref{equation:ipt_zero}, using the score in~\autoref{equation:ipt_qwen_score}.
The language-modeling loss is computed over all training instances, irrespective of whether they participate in an intervention pair.
Thus, instances with available pairs receive additional intervention-level supervision, whereas unpaired instances receive only the standard language-modeling objective and reduce to ordinary SFT.

\subsection{GEO Detection Evaluation}
\label{section:binary_eval}

\begin{table}[!t]
\centering
\caption{Comparison of selected GEO detectors and our proposed methods.
WGA denotes worst-group accuracy, while $\Delta$FPR and $\Delta$TPR measure
absolute error-rate gaps between AI- and human-authored groups.}
\label{table:ipt_comparison}
\setlength{\tabcolsep}{1.5pt}
\renewcommand{\arraystretch}{1.05}
\customTableFont

\begin{tabular}{@{}lccccc@{}}
\toprule
& \multicolumn{2}{c}{\textbf{Overall Performance}}
& \multicolumn{3}{c}{\textbf{Shortcut Diagnostics}} \\
\cmidrule(lr){2-3}
\cmidrule(l){4-6}
\textbf{Method}
& \textbf{Acc.}
& \textbf{F1}
& \textbf{WGA$\uparrow$}
& \textbf{$\Delta$FPR$\downarrow$}
& \textbf{$\Delta$TPR$\downarrow$} \\
\midrule

\multicolumn{6}{@{}l}{\textbf{\textit{Selected baselines}}} \\
ModernBERT (1K)
& 0.842 & 0.864 & 0.725 & 0.271 & 0.121 \\
ModernBERT (8K)
& 0.839 & 0.862 & 0.725 & 0.263 & 0.117 \\
Qwen Zero-shot
& 0.627 & 0.770 & 0.008 & 0.000 & 0.006 \\
Qwen SFT
& 0.778 & 0.833 & 0.325 & 0.413 & 0.105 \\
TF-IDF char (2--4) + LR
& 0.848 & 0.878 & 0.575 & 0.346 & 0.077 \\
TF-IDF word (1--2) + LR
& 0.846 & 0.880 & 0.375 & 0.558 & 0.111 \\
Pangram
& 0.838 & 0.876 & 0.192 & 0.767 & 0.099 \\

\midrule
\multicolumn{6}{@{}l}{\textbf{\textit{Fine-tuned models with GRL/IPT}}} \\
ModernBERT-GRL
& 0.876 & 0.897 & 0.783 & 0.171 & 0.069 \\
ModernBERT-IPT
& 0.931 & 0.944 & 0.883 & 0.108 & 0.062 \\
Qwen-IPT
& 0.860 & 0.883 & 0.775 & 0.158 & 0.033 \\

\bottomrule
\end{tabular}
\end{table}

\mypara{Experimental Settings}
All experiments use the same fixed query-disjoint split described in~\autoref{section:geoflagbench_setting}, containing 2,238 training instances and 962 test instances.
The detection threshold is $0.5$.
We fine-tune ModernBERT-base and Qwen3-0.6B.
For all models, we empirically use a learning rate of $2\times10^{-5}$, margin $m=2$, $\alpha=1$, and $\beta=1$.
For ModernBERT-base, we use an 8,192-token context and 5 epochs, and for Qwen3-0.6B, we use a 4,096-token context, 10 epochs, and LoRA rank 16.

\mypara{Results}
IPT improves both overall performance and shortcut diagnostic metrics.
For ModernBERT, IPT increases accuracy from 0.839 to 0.931 and F1 from 0.862 to 0.944.
The improvement is also clear in the shortcut diagnostics: WGA rises from 0.725 to 0.883, while $\Delta$FPR and $\Delta$TPR decrease from 0.263 and 0.117 to 0.108 and 0.062, respectively.
Qwen does not outperform ModernBERT, but when using IPT, the performance also improves.
Compared with Qwen-SFT, Qwen-IPT improves accuracy from 0.778 to 0.860 and F1 from 0.833 to 0.883, while increasing WGA substantially from 0.325 to 0.775.
At the same time, $\Delta$FPR drops from 0.413 to 0.158 and $\Delta$TPR from 0.105 to 0.033.
Qwen Zero-shot has very small $\Delta$FPR and $\Delta$TPR, but its WGA is only 0.008, indicating that the small group gaps mainly result from poor classification performance rather than genuine robustness to authorship shortcuts.
GRL shows the same overall trend, but the gains are smaller.
ModernBERT-GRL reaches 0.876 accuracy and 0.897 F1, improves WGA to 0.783, and reduces $\Delta$FPR and $\Delta$TPR to 0.171 and 0.069.
Overall, the results show that IPT improves performance across different GEO methods and authorship conditions without disproportionately favoring any single group.

\subsection{GEO Attribution Evaluation}
\label{section:attribution}

\begin{table}[!t]
\centering
\caption{Results on the seven-class attribution task.}
\label{table:attribution_all_results}
\setlength{\tabcolsep}{2pt}
\customTableFont
\begin{tabular}{lrr}
\toprule
\textbf{Method} & \textbf{Acc.} & \textbf{Macro F1} \\
\midrule
TF-IDF Character 2--4 with LR        & 0.730 & 0.612 \\
TF-IDF Word 1--2 with LR             & 0.710 & 0.562 \\
Ghostbuster features, recalibrated   & 0.545 & 0.286 \\
Structural features with XGBoost     & 0.648 & 0.573 \\
GPT-2 perplexity features with LR    & 0.497 & 0.226 \\
Gemini-3.5-flash                     & 0.500 & 0.315 \\
Haiku-4.5                            & 0.402 & 0.314 \\
ModernBERT SFT                       & 0.830 & 0.773 \\
ModernBERT IPT                       & 0.906 & 0.895 \\
\bottomrule
\end{tabular}
\end{table}

\mypara{Task Description}
We further evaluate whether IPT extends beyond binary GEO detection by
considering a multiclass GEO attribution task.
Given the text of a webpage, the model predicts one of seven classes:
\textit{non-GEO}, \textit{AutoGEO}, \textit{GEO Strategy Pool},
\textit{PMA}, \textit{RAID}, \textit{Meta-Optimization}, or
\textit{Human GEO}.
Stealthy AutoGEO is grouped with \textit{AutoGEO}, and Stealthy GEO Strategy
Pool is grouped with \textit{GEO Strategy Pool}.
This setting allows us to examine whether the intervention-based training
objective of IPT remains effective when the prediction target additionally
distinguishes among different GEO method families.

\mypara{IPT for Attribution}
We adapt IPT to the seven-class attribution task while preserving its
intervention-based constraints.
Let $\ell_c(x)$ denote the logit for class $c$.
To apply the pairwise IPT objectives, we aggregate the attribution logits into
a binary GEO score:
\begin{equation}
z_{\theta}(x)
=
\log\!\sum_{c\in\mathcal{C}_{\mathrm{GEO}}}
\exp\!\left(\ell_c(x)\right)
-
\ell_{\mathrm{non\text{-}GEO}}(x),
\label{equation:ipt_attribution_score}
\end{equation}
where $\mathcal{C}_{\mathrm{GEO}}$ contains the six GEO attribution classes.
The standard classification loss is replaced by seven-class cross-entropy,
while the positive and zero pair constraints are applied to
$z_{\theta}(x)$ as in the binary setting.

\mypara{Experiment Settings}
We use the same query-level training and test split as in
\autoref{section:current_methods}.
The baseline methods follow the same preprocessing procedures, model
configurations, context lengths, feature extraction settings, and training
hyperparameters described in~\autoref{section:current_methods}.
For supervised classifiers, we replace the binary target with the seven
attribution classes while keeping the remaining configuration unchanged.
For IPT, we use the same settings in~\autoref{section:binary_eval}.
We report macro F1 and accuracy as the primary attribution metrics.

\mypara{Results}
On the seven-class attribution task, IPT substantially improves the model's ability to identify which GEO method produced a given instance.
As shown in \autoref{table:attribution_all_results}, ModernBERT IPT achieves a Macro F1 of 0.895 and an accuracy of 0.906, improving over the same ModernBERT backbone with SFT by 0.122 and 0.076, respectively, while also clearly outperforming the feature-based and general-purpose LLM baselines.
Importantly, this gain is reflected in more balanced per-class performance.
As shown in \autoref{table:attribution_all_recall}, SFT already performs well on Non-GEO, AutoGEO, and Pool, but remains considerably weaker on RAID, Meta, and Human, with recall on the Human class reaching only 0.333.
IPT raises the recall of RAID, Meta, and Human to 0.909, 0.917, and 0.759, respectively, while maintaining recall between 0.855 and 0.906 on AutoGEO, Pool, and PMA.
The per-class precision results in \autoref{table:attribution_all_precision} show a similar pattern: IPT achieves at least 0.804 precision for every class, including 0.985 for PMA and 0.984 for RAID.
At the same time, these improvements do not come at the expense of distinguishing non-GEO instances, for which IPT retains a recall of 0.950 and a precision of 0.891.
Overall, the results suggest that IPT learns more than a coarse distinction between GEO and Non-GEO content; it also captures discriminative signals associated with different GEO optimization mechanisms, enabling substantially stronger fine-grained attribution.

\section{A GEO-Gated Agent System for Assessing Citation URL Verifiability}
\label{section:extension}

\subsection{Motivation}
\label{section:extension_scope}

GEO methods may be misused to make weak or false information appear well supported in the real world.
For example, an operator can first publish a false claim on a website that is easy to create or edit, and then cite that page from another website~\cite{fan2026geo,cna2026geo}.
The citation URL is valid, and the corresponding page may directly support the claim, but the source itself was \emph{strategically} created to support it.

\subsection{Audit Scope and Procedure}
\label{section:audit_intro}

\begin{figure}[!t]
\centering
\includegraphics[width=.96\columnwidth]{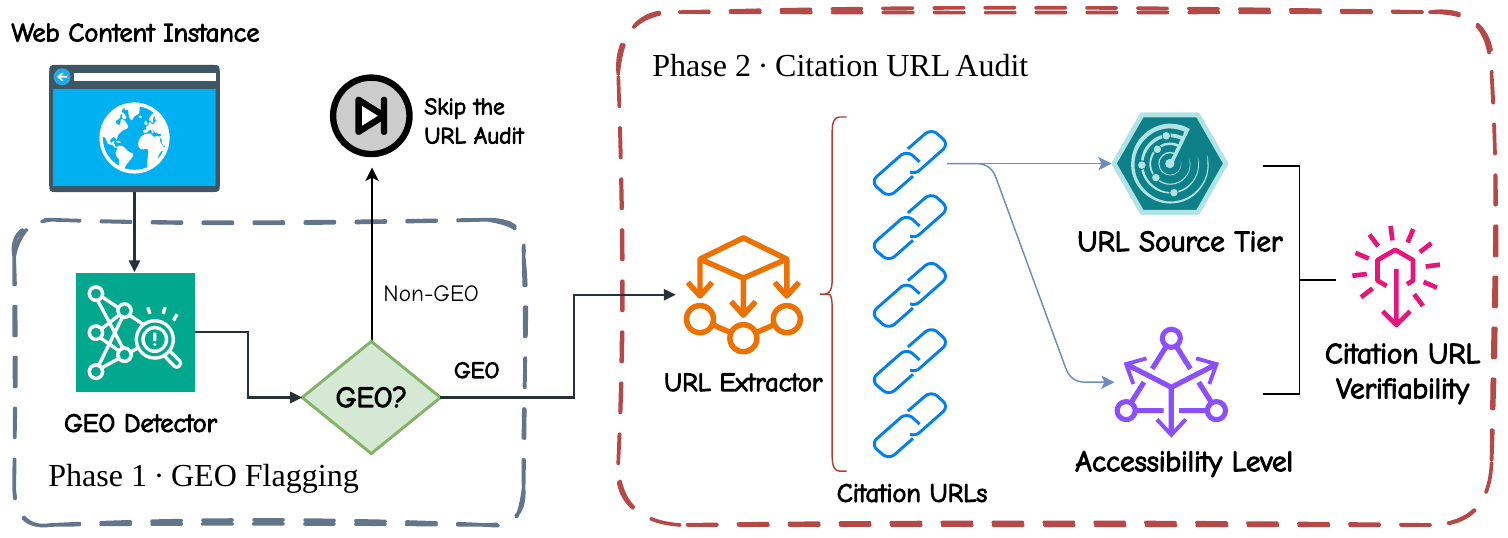}
\caption{Overview of the Two-Phase Audit.}
\label{figure:two_phase}
\end{figure}

To address this issue, we define a two-phase audit task that operates at both the web-content-instance level and the citation-URL level, as outlined in~\autoref{figure:two_phase}.

\mypara{Phase 1: GEO Identification}
\emph{Given a web content instance, determine whether it is GEO-optimized.}
This phase produces a web-content-instance-level label, either \textsc{GEO} or \textsc{Non-GEO}.
Instances labeled as \textsc{Non-GEO} are not audited further.
Instances labeled as \textsc{GEO} proceed to the second phase, where their citation URLs are audited.

\mypara{Phase 2: Citation URL Audit}
\emph{For an instance identified as GEO-optimized, we audit the verifiability of each citation URL contained in the content.}
Here, citation URL verifiability considers two factors: the accountability and editorial control of the cited source, and the accessibility of the cited content.\footnote{Citation URL verifiability characterizes the inspectability and source accountability of a citation, rather than the factual correctness of the cited claim or whether the cited content semantically supports it.}

We operationalize these two factors as follows.
First, \emph{URL source tier} captures the accountability and editorial control of the publisher behind a citation URL.
Following previous studies~\cite{JZXQZD26,LS26}, we define three tiers based on the source's publication and review process: \textsc{C1} (primary or strongly accountable authority), \textsc{C2} (accountable secondary or curated source), and \textsc{C3} (limited-accountability or user-contributed source), in decreasing order of accountability and editorial control.
The complete definition is given in~\autoref{table:citation_url_source_tier_definition}.
Second, \emph{accessibility} captures whether the cited URL can be visited.
We define three accessibility levels: directly accessible, restricted or archived, and unavailable or unresolved.
The full definition is in~\autoref{table:citation_url_accessibility_definition}.
We treat accessibility separately because even an accountable source may have low verifiability if its cited content cannot be accessed.

\begin{figure}[!t]
\centering
\includegraphics[width=0.56\columnwidth]{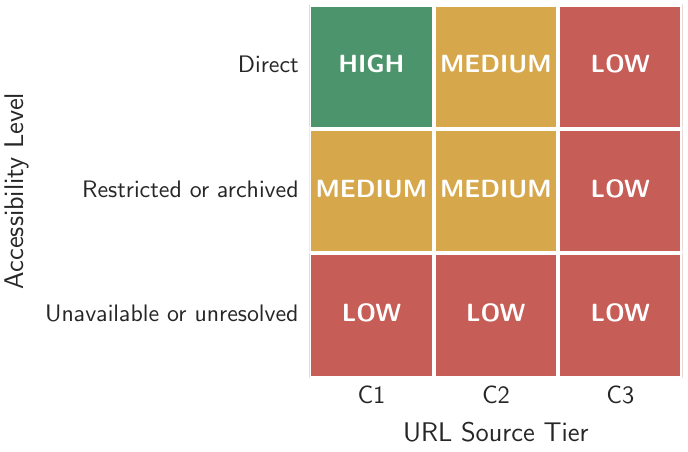}
\caption{Deterministic mapping from URL source tier and accessibility level to citation URL verifiability.
}
\label{figure:citation_url_verifiability_definition}
\end{figure}

For each citation URL occurrence, URL source tier and accessibility are combined according to a predefined mapping to produce the final \emph{citation URL verifiability} label: \textsc{High}, \textsc{Medium}, or \textsc{Low}.
In general, citations from higher-tier sources receive higher verifiability labels when the cited content is accessible, whereas limited source accountability or failed retrieval lowers the resulting label.
The complete mapping from URL source tier and accessibility to the three verifiability levels is given in~\autoref{figure:citation_url_verifiability_definition}.
Accordingly, Phase~2 outputs one verifiability label for each citation occurrence in every GEO-identified instance.

\subsection{Test Dataset}
\label{section:extension_benchmark}

\mypara{Dataset Construction}
The test set contains \emph{562 web content instances}, including 281 non-GEO instances and 281 corresponding GEO-optimized instances.
They are constructed from 281 queries across the same four domains: 54 in Finance, 68 in Health, 97 in Technology, and 62 in Travel.
These queries do not overlap with those used in \texttt{GEOFlagBench}.
For each query, we collect one non-GEO instance and generate one corresponding GEO instance following the procedure in~\autoref{section:geoflagbench}.
As in \texttt{GEOFlagBench}, the non-GEO instances are collected from webpage snapshots dated before October 2022.
Specifically, among the 281 GEO instances described above, 129 contain at least one citation, while the remaining 152 contain no citations.
The 129 citation-bearing GEO instances contain 2{,}160 citation occurrences in total.

\mypara{Citation URL Annotation}
For the above 2{,}160  citation URL occurrences, the values of citation URL verifiability are determined along two dimensions: \emph{URL source tier} and \emph{accessibility}.

For \emph{URL source tier}, two annotators independently assign each citation URL occurrence to \textsc{C1}, \textsc{C2}, or \textsc{C3} according to~\autoref{table:citation_url_source_tier_definition}.
They do not see each other's annotations or the GEO method used to generate the corresponding web content instance.
A third annotator reviews all disagreements, and the final URL source tier label is determined by majority vote.
The two primary annotators achieve 80.39\% agreement, with Cohen's $\kappa=0.70$.

For \emph{accessibility}, we use a deterministic retrieval procedure rather than human annotation.
For each citation URL, the retriever performs an HTTP GET with redirects, a 20-second timeout, and up to two attempts.
If live retrieval fails, it additionally queries the Internet Archive for an available snapshot.
The retrieval may return six potential values, and the returned values are mapped to three accessibility levels according to~\autoref{table:citation_url_accessibility_definition}.

Finally, the URL source tier and accessibility level are combined according to the predefined mapping in~\autoref{figure:citation_url_verifiability_definition} to derive the \emph{citation URL verifiability} label.
As this label depends on the manually annotated source tier, we also measure how source-tier disagreement propagates to verifiability.
The resulting agreement on citation URL verifiability is 84.16\%, with Cohen's $\kappa=0.75$.

\subsection{Automatic System Design}
\label{section:extension_system_design}

We propose an automatic auditing system for the two-phase audit task:
A top-level \emph{audit agent} orchestrates five tools to (1) determine the \emph{GEO Flag}, (2) extract citation URLs, (3) assess each citation URL along the two verifiability dimensions, and (4) derive the final Citation URL verifiability label.
The five tools share a common interface but may execute deterministic procedures or invoke specialized subagents, as described below.

\mypara{GEO Flag Tool}
This tool implements Phase~1.
It applies a GEO detector to the input web content instance and returns its GEO prediction.
A \textsc{Non-GEO} prediction terminates the audit, whereas a \textsc{GEO} prediction triggers citation URL processing.

\mypara{Citation URL Extraction Tool}
For each web content instance that passes the GEO Flag tool, the tool identifies citation URL occurrences and normalizes their destination URLs.
It is implemented as a deterministic parser that resolves relative URLs, canonicalizes URLs, and removes known tracking parameters.
If no citation URL is found, the audit terminates with \texttt{NO\_CITATION}.

\mypara{Citation URL Accessibility Tool}
For each extracted citation URL, the tool assesses the accessibility dimension.
It is backed by a specialized accessibility subagent, which invokes the same deterministic retriever described in~\autoref{section:extension_benchmark} and returns the resulting accessibility level.

\mypara{Citation URL Source Tier Tool}
In parallel, the tool assesses the URL source tier dimension.
It is backed by a specialized source-tier subagent.
Given the citation URL, the subagent assigns \textsc{C1}, \textsc{C2}, or \textsc{C3} according to~\autoref{table:citation_url_source_tier_definition}.
The implementation additionally supports web search and source-list lookup, but both are disabled in our experiments, and the source lists are empty.

\mypara{Citation URL Verifiability Aggregation Tool}
The tool deterministically derives the final citation URL verifiability label from the URL source tier and accessibility level according to~\autoref{figure:citation_url_verifiability_definition}.

The resulting audit contains one \emph{GEO Flag} for the web content instance and, for each extracted citation URL occurrence, its \emph{URL source tier} and \emph{accessibility level} together with the \emph{citation URL verifiability} label derived from these two dimensions.

\subsection{Evaluation}
\label{section:extension_evaluation}

\begin{table}[!t]
\centering
\caption{Conditional performance on 1,653 citation occurrences from gold-labeled GEO instances correctly detected by the GEO gate.
Each cell reports accuracy / macro F1 (\%).}
\label{table:extension_agent_models}
\setlength{\tabcolsep}{3pt}
\customTableFont
\begin{tabular}{lcc}
\toprule
\textbf{Agent Model}
& \textbf{URL Source Tier}
& \textbf{Citation URL Verifiability} \\
\midrule
GLM 5.2             & \textbf{84.75 / 84.02} & \textbf{83.00 / 83.20} \\
Kimi K3             & 78.46 / 76.43          & 76.65 / 76.78 \\
Qwen 3.6 35B-A3B    & 77.37 / 75.69          & 76.10 / 76.54 \\
GPT 5.6 Luna        & 76.77 / 75.03          & 75.32 / 75.50 \\
Claude Haiku 4.5    & 73.93 / 71.39          & 72.60 / 72.50 \\
Gemini 3.5 Flash    & 62.07 / 58.10          & 62.31 / 63.05 \\
\bottomrule
\end{tabular}
\end{table}

\mypara{Experimental Settings}
We use the ModernBERT-IPT classifier from~\autoref{section:ipt} as the GEO detector in the GEO Flag Tool.
For the agent-based components, we evaluate six frontier LLMs: Kimi K3, Qwen 3.6 35B-A3B, GLM 5.2, Claude Haiku 4.5, GPT 5.6 Luna, and Gemini 3.5 Flash.
We set the temperature to zero when the model API supports this option; otherwise, we use the default generation settings.

We evaluate the two audit phases separately.
For GEO flagging, we report instance-level accuracy and F1, treating \textsc{GEO} as the positive class.
For citation URL auditing, we report accuracy and macro F1 for both URL source tier and citation URL verifiability.

\mypara{GEO Flagging Performance}
We first evaluate Phase~1 at the \emph{web-content-instance level} on all 562 web content instances, including 281 GEO and 281 non-GEO instances.
The GEO Flag Tool achieves 93.95\% accuracy and 93.63\% F1, with 98.81\% precision and 88.97\% recall for the GEO class.

\mypara{Citation URL Labeling Performance}
We evaluate Phase~2 at the \emph{citation-URL level} on 1{,}653 citation URL occurrences from gold-labeled GEO instances that are correctly identified by the GEO Flag Tool.\footnote{
Among the 129 gold-labeled GEO instances containing citations, 106 are correctly identified as GEO and proceed to citation URL auditing.
These 106 instances contain 1{,}653 of the 2{,}160 citation URL occurrences, corresponding to 82.17\% instance coverage and 76.53\% citation URL occurrence coverage.
Three additional non-GEO instances also pass the GEO Flag Tool but are excluded from this conditional evaluation.
}
This conditional evaluation isolates the citation URL labeling capability after a gold-labeled GEO instance passes the GEO Flag Tool.
\autoref{table:extension_agent_models} reports accuracy and macro F1 for both URL source tier and citation URL verifiability.

GLM 5.2 achieves the strongest performance on both outputs.
For URL source tier, it reaches 84.75\% accuracy and 84.02\% macro F1.
For citation URL verifiability, it reaches 83.00\% accuracy and 83.20\% macro F1.
Compared with the second-best model, GLM 5.2 improves accuracy by 6.29 percentage points for URL source tier and 6.35 percentage points for citation URL verifiability.
Looking at per-class performance, GLM 5.2 achieves recalls of 94.77\%, 85.05\%, and 71.46\% for \textsc{C1}, \textsc{C2}, and \textsc{C3}, respectively, with corresponding precision values of 88.46\%, 76.86\%, and 90.39\%.
For citation URL verifiability, recall is 90.27\% for \textsc{High}, 84.67\% for \textsc{Medium}, and 75.84\% for \textsc{Low}, with corresponding precision values of 79.04\%, 79.45\%, and 92.47\%.
The high precision for LOW similarly indicates that LOW predictions are highly reliable when produced.
In addition, \autoref{table:extension_agent_efficiency} in \refappendix{section:extra_tab_fig} reports the runtime and estimated inference cost.
GLM 5.2 achieves the best labeling performance with a moderate runtime of 29.92 minutes and an estimated cost of \$3.72.

Overall, the above results show that the audit system (driven by GLM 5.2) provides relatively strong performance.

\section{Empirically Estimating GEO Prevalence in Real-World Search Results}
\label{section:real_world_audit}

Using the automatic audit system developed in~\autoref{section:extension}, we conduct a real-world audit of pages linked by conventional Google Search and Gemini-grounded search.
We estimate the prevalence of GEO signals across the two retrieval channels.
We also examine how the detection rates vary with declared modification time and publishing environment.
For detected GEO pages, we further audit the URL source tier and citation URL verifiability labels.\footnote{We report these results as estimates rather than definitive measurements for two reasons.
The live pages lack groundtruth GEO labels.
The GEO detector and the citation URL audit pipeline can also make prediction errors.}

\subsection{Experimental Settings}

\mypara{Query Sampling and Retrieval Channels}
We use the 5,000-query ORCAS dataset~\cite{CCMYB20,GLCSBC26} which contains clicked query and document pairs collected from real users.

We normalize each query by case folding, trimming surrounding spaces, and collapsing repeated spaces.
This process identifies one duplicate and leaves 4,999 eligible queries.
We sample 1,000 queries through proportional stratified random sampling~\cite{C77}.
Each stratum combines an upstream intent label with a query length bin.
The intent labels are Factual, Instrumental, Navigational, Transactional, and Abstain.
The length bins contain one, two, three, four, five, or at least six query terms.
We assign integer stratum quotas with the Hamilton largest-remainder method~\cite{BY01}.
The selection does not use released URLs, result ranks, retrieval outcomes, or GEO predictions.

For each selected query, we use the Google Search and Gemini URL lists released by Grossman et al.~\cite{GLCSBC26}.
The Google Search channel contains landing-page links from the conventional Google Search results page~\cite{GLCSBC26,serpapiGoogleSearchApi}.
The Gemini channel contains source links attached to Gemini 2.5 Flash answers generated with Google Search grounding~\cite{GLCSBC26,googleGrounding}.
Google Search grounding allows Gemini to retrieve web sources and attach those sources to its answer.

Google Search provides at least one released URL for all 1,000 sampled queries.
Gemini provides at least one released URL for 979 queries.
The other 21 Gemini records contain no released URL.
The release does not distinguish an answer without source links from a collection failure.
We retain all 1,000 queries in the Google Search analysis and use the 979 queries with released links in the Gemini analysis.
We do not treat an empty Gemini record as a query with a zero GEO rate.
The two channels contain 16,568 URL occurrences in total.
Google Search contributes 7,856 unique URLs, and Gemini contributes 8,591 unique URLs.
Their union contains 13,985 unique URLs because some pages appear in both channels.
We preserve every query, channel, rank, and URL association for exposure and paired-query analyses.
We assign one identifier to each normalized URL so that a shared page is fetched and analyzed once.

\mypara{Page Collection and Recovery}
We fetched the current page versions from 28 to 31 July 2026.
The initial deterministic fetcher followed redirects, respected robots policies, recorded HTTP outcomes, and extracted the main text as Markdown.
An extraction was initially considered usable when it contained at least 300 characters of page text.

We applied several recovery stages to pages without usable text.
The first stage rendered eligible pages in Chrome and blocked images, media, and fonts.
The second stage retried transient failures and extracted text from PDF files.
The third stage targeted unresolved navigation and extraction failures.
The final stage reviewed every remaining extraction below 300 characters and retained only pages with meaningful page content.
We define meaningful content as a coherent representation of the requested public page that agrees with its URL or title and contains page-specific facts or usable functionality.
Navigation-only text, cookie notices, login or payment forms, access challenges, error messages, loading shells, unrelated fragments, and isolated metadata are not meaningful page content.
Text length alone did not determine this decision, so a complete and identifiable public page below 300 characters could be retained.
We did not bypass robots policies, authentication, paywalls, CAPTCHAs, or terminal HTTP errors.

The initial and browser stages retained 7,896 unique pages.
Broad recovery added 2,063 pages.
After the initial and browser stages, 6,089 URLs had no usable content.
Broad recovery attempted 5,775 of these URLs and retained 2,063, which gives a recovery rate of 35.72\% among attempted URLs.
The other 314 URLs had terminal or policy outcomes that were outside the permitted recovery procedure.
Targeted recovery added 110 pages.
Targeted recovery attempted 529 unresolved URLs and retained 110, which gives a recovery rate of 20.79\%.
The short-content review added 26 pages.
The short-content stage reviewed all 171 remaining candidates below 300 characters in the two-channel URL set.
It retained 22 recovered pages and four complete pages from the original extraction, for a total of 26 meaningful pages and a retention rate of 15.20\%.
The final corpus therefore contains 10,095 usable pages from 13,985 unique URLs, which gives 72.18\% retrieval coverage.
The channel-specific and union coverage values are summarized in~\autoref{table:real_world_coverage} of~\refappendix{section:real_world_audit_pipeline}.
The remaining 3,890 URLs did not yield usable page content.
We assign each unavailable URL to its last recorded outcome after all permitted recovery stages and report them in~\autoref{table:real_world_unavailable}.

\mypara{Pipeline Configuration}
We apply the frozen pipeline from~\autoref{section:extension} without retraining or audit-specific calibration.
Every usable page first passes through the GEO detector.
Only pages detected as GEO enter the deterministic citation parser and the GLM 5.2 citation URL audit.
We use the URL source tier and citation URL verifiability definitions from~\autoref{section:extension}.

\mypara{Analysis Units and Statistical Procedures}
The primary page-level estimate uses the 10,095 unique pages in the channel union.
A shared page contributes once to this estimate.
Each channel-specific page-level GEO estimate in~\autoref{table:real_world_geo_comparison} includes all usable pages returned by that channel.
A shared page contributes once to each channel that returned it.

Among the 979 queries with at least one released URL from each channel, 965 yield at least one usable page in each channel and enter the paired analysis.
For the remaining 14 queries, at least one channel yields no usable page: eight for Google Search and six for Gemini.
For each retained query, we compute the fraction of usable pages detected as GEO in each channel.
We then average these fractions across queries.
We compute the 95\% interval for the difference with 10,000 paired bootstrap samples.

For the \emph{modification-time analysis}, the unit is one usable page.
We group a page by year only when its metadata contains a parseable \texttt{dateModified} value.
A page enters the missing group when this field is absent or invalid.
We do not replace it with \texttt{datePublished}, because publication time and modification time describe different events.

For the \emph{citation URL analysis}, the unit is one citation occurrence extracted from a page detected as GEO.
The same citation URL is counted again when it appears as another citation occurrence on the same or another page.
For channel-specific results, a citation on a page returned by both channels contributes once to the Google Search row and once to the Gemini row.
The unique-union result includes that shared page once and therefore counts the citation once.

The audit progress and the distinction between Page URLs returned by search systems and citation URLs embedded in the fetched page content are summarized in~\autoref {figure:real_world_audit_pipeline} of~\refappendix{section:real_world_audit_pipeline}.

\subsection{Estimated GEO Prevalence}

\mypara{Retrieval Channels}
The pipeline detects 898 of the 10,095 unique pages as GEO, corresponding to an overall prevalence estimate of 8.90\% (95\% Wilson CI~\cite{W27}: [8.36\%, 9.47\%]).
However, the aggregate estimate masks a systematic difference between retrieval channels.
As shown in~\autoref{table:real_world_geo_comparison}, GEO is detected in 8.14\% of Google Search pages and 9.09\% of Gemini pages, a page-level difference of 0.95 pp.

This difference persists and becomes larger after accounting for unequal numbers of retrieved pages across queries.
For each of the 965 queries with at least one usable page from both channels, we compute the within-query GEO fraction for each channel and then average these fractions across queries, giving every query equal weight.
The resulting macro rates are 7.91\% for Google Search and 9.34\% for Gemini, corresponding to a Gemini--Google difference of 1.43 percentage points (paired-bootstrap 95\% CI: [0.46, 2.41] pp).

These results indicate that GEO is unevenly represented across retrieval channels, with Gemini exhibiting consistently higher GEO prevalence than Google Search.
The persistence of this difference under the query-balanced comparison suggests that the two channels are affected by GEO to different degrees, rather than the observed gap being driven solely by differences in retrieval volume across queries.

\begin{table*}[!ht]
\centering
\caption{GEO estimates and comparisons between retrieval channels.
An example is available in~\autoref{figure:paired_query_picot} in~\refappendix{section:extra_tab_fig}.
}
\label{table:real_world_geo_comparison}
\setlength{\tabcolsep}{6pt}
\customTableFont
\begin{tabular}{@{}lccccc@{}}
\toprule
\textbf{Setting} & \textbf{Google Search} & \textbf{Gemini} & \textbf{Unique Union} & \textbf{Gemini $-$ Google} & \textbf{95\% CI} \\
\midrule
Page-level estimate & 441/5,421 (8.14\%) & 599/6,590 (9.09\%) & 898/10,095 (8.90\%) & 0.95 pp & N/A \\
Paired-query macro ($n=965$) & 7.91\% & 9.34\% & N/A & 1.43 pp & [0.46, 2.41] pp \\
\bottomrule
\end{tabular}
\end{table*}

\begin{figure}[!t]
\centering
\begin{subfigure}{0.48\columnwidth}
\centering
\includegraphics[width=1\linewidth]{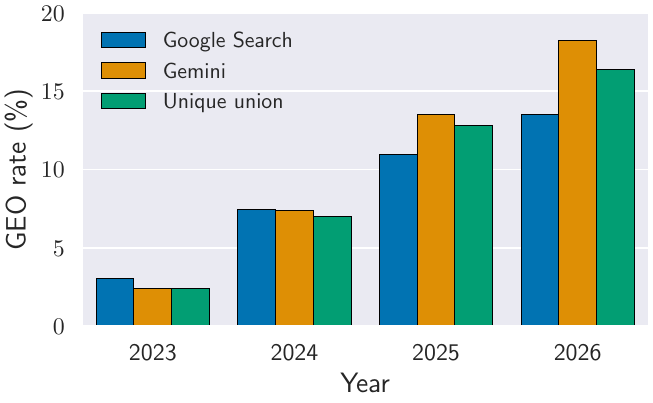}
\caption{GEO Detection Rate}
\label{figure:real_world_time_rate}
\end{subfigure}
\hfill
\begin{subfigure}{0.48\columnwidth}
\centering
\includegraphics[width=1\linewidth]{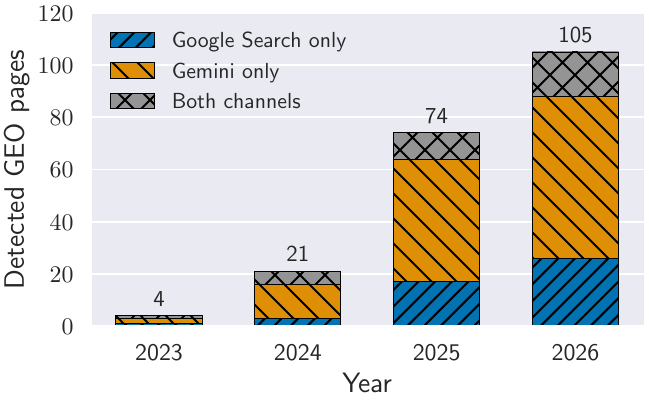}
\caption{Detected GEO Page Count}
\label{figure:real_world_time_count}
\end{subfigure}
\caption{GEO prevalence by declared \texttt{dateModified} year (2023--2026).
Panel (a) reports detection rates for Google Search, Gemini, and their unique union; Panel (b) partitions union GEO detections into three mutually exclusive exposure groups.
Pages without a parseable \texttt{dateModified} are omitted: 8,119 union pages (682 GEO; 8.40\%), 4,563 Google pages (355; 7.78\%), and 5,153 Gemini pages (434; 8.42\%).
}
\label{figure:real_world_time}
\end{figure}

\mypara{Modification Time}
A parseable \texttt{dateModified} value is available for 1,976 of the 10,095 unique pages, corresponding to 19.57\% metadata coverage.
We study those pages modified between 2023 and 2026 (1,684 pages).
As shown in~\autoref{figure:real_world_time}, estimated GEO prevalence is relatively low among pages modified in 2023, but increases markedly from 2024 onward.
In the unique union, the estimated rate rises from 7.02\% in 2024 to 12.80\% in 2025 and 16.36\% in 2026.
The same temporal pattern appears in both retrieval channels, with 2026 estimates reaching 13.52\% for Google Search and 18.20\% for Gemini.
These results suggest that GEO is more prevalent among recently modified pages.

\mypara{Domain-Level Variation}
To examine domain-level variation, we report all domains with more than 50 successfully retrieved usable pages in the unique-page union.
As shown in~\autoref{table:real_world_domains}, estimated GEO prevalence varies substantially across these frequently retrieved domains.
Several large informational and health-related domains contain no detected GEO pages in our sample, whereas YouTube reaches 4.38\% and Amazon reaches 20.37\%.
We present two examples (from YouTube and Amazon) in~\autoref{table:geo_cases} in~\refappendix{section:extra_tab_fig}.
One possible explanation is that commercial and creator-driven platforms provide stronger incentives to optimize content for visibility in AI-mediated discovery.
YouTube has introduced conversational search for video discovery alongside Gemini-powered tools for content creation and remixing~\cite{youtube_conversational_search_2026}.
For example, Amazon explicitly advises sellers that understanding how its generative shopping assistant interprets product information can help them create more effective listings~\cite{amazon_rufus_2024,amazon_rufus_sellers_2026,amazon_search_listings_2026}, while YouTube has introduced Gemini-powered conversational search for video discovery~\cite{youtube_conversational_search_2026}.
While our data cannot establish the intent behind individual pages, these developments provide a plausible explanation for why GEO may be more prevalent on commercially or visibility-driven platforms.

\begin{table}[!t]
\centering
\caption{Largest domains in the unique-page union.}
\label{table:real_world_domains}
\setlength{\tabcolsep}{3pt}
\customTableFont
\begin{tabular}{lrrr}
\toprule
\textbf{Domain} & \textbf{Pages} & \textbf{GEO} & \textbf{GEO Rate} \\
\midrule
\texttt{en.wikipedia.org} & 613 & 0 & 0.00\% \\
\texttt{youtube.com} & 434 & 19 & 4.38\% \\
\texttt{my.clevelandclinic.org} & 106 & 0 & 0.00\% \\
\texttt{webmd.com} & 78 & 0 & 0.00\% \\
\texttt{healthline.com} & 73 & 0 & 0.00\% \\
\texttt{medicalnewstoday.com} & 61 & 2 & 3.28\% \\
\texttt{pmc.ncbi.nlm.nih.gov} & 58 & 0 & 0.00\% \\
\texttt{amazon.com} & 54 & 11 & 20.37\% \\
\texttt{nhs.uk} & 53 & 0 & 0.00\% \\
\bottomrule
\end{tabular}
\end{table}

\begin{tcolorbox}[
title=Takeaway,
colback=cyan!3,
colframe=cyan!45!black,
fontupper=\small,
fonttitle=\bfseries\small,
boxrule=0.6pt,
arc=1mm,
before skip=8pt,
after skip=8pt]
Within the data used in our estimation, more than 8\% of pages returned by both conventional Google Search and Gemini-grounded search are estimated to exhibit GEO, with a higher estimated prevalence for Gemini-grounded search.
In addition, both the count and proportion of pages estimated to exhibit GEO have increased year by year since 2024.
\end{tcolorbox}

\subsection{Estimated Verifiability of Citation URLs}

This analysis applies to the 898 pages detected as GEO in the preceding analysis.
Among them, 489 contain at least one parsed citation.
The other 409 pages contain no parsed citation and do not enter the citation URL analysis.
The unique union contains 6,663 citation occurrences.
All subsequent tables in this subsection use citation occurrences as the analysis unit, so repeated uses of the same citation URL are counted separately.
Definitions of verifiability-related metrics are introduced in~\autoref{section:audit_intro}.\footnote{This measure does not assess whether a cited claim is true or whether the content at the citation URL supports it.}

\begin{figure}[!t]
\centering
\begin{subfigure}{0.48\columnwidth}
\centering
\includegraphics[width=.96\linewidth]{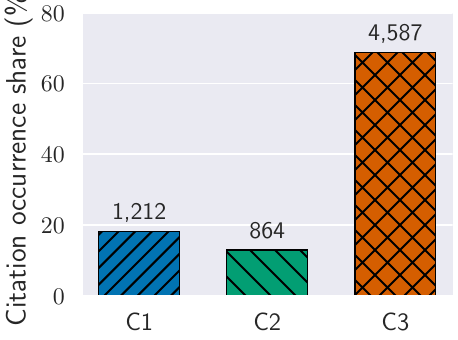}
\caption{URL source tier}
\label{figure:citation_url_source_tier}
\end{subfigure}
\hfill
\begin{subfigure}{0.48\columnwidth}
\centering
\includegraphics[width=.96\linewidth]{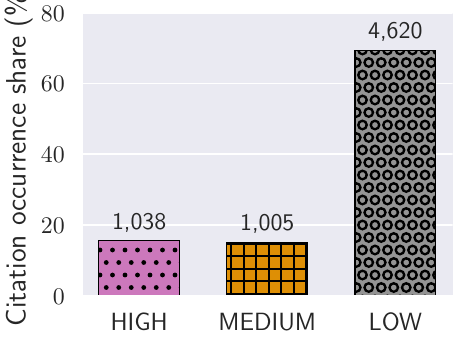}
\caption{Citation URL verifiability}
\label{figure:citation_url_verifiability}
\end{subfigure}
\caption{Citation URL labels for 6,663 occurrences in the unique-page union.
Bars show shares, with occurrence counts above them.
The two panels are separate marginal distributions and do not imply one-to-one category mappings.}
\label{figure:real_world_citation_url_labels}
\end{figure}

\begin{table}[!t]
\centering
\caption{Page and citation URL statistics by retrieval channel.
The channels overlap in 142 GEO pages, including 82 cited pages and 771 citation occurrences.
Thus, the unique union contains 898 GEO pages and 6{,}663 citation occurrences.}
\label{table:real_world_citation_url_channel}
\setlength{\tabcolsep}{3pt}
\customTableFont
\begin{tabular}{lrr}
\toprule
\textbf{Metric} & \textbf{Google Search} & \textbf{Gemini} \\
\midrule
\multicolumn{3}{l}{\textbf{\textit{Page and Citation Volume}}} \\
Detected GEO pages & 441 & 599 \\
GEO pages with citations & 236 & 335 \\
GEO pages with citations (\% of GEO pages) & 53.51\% & 55.93\% \\
Citation occurrences & 2,162 & 5,272 \\
Citations per detected GEO page & 4.90 & 8.80 \\
Citations per cited GEO page & 9.16 & 15.74 \\
\midrule
\multicolumn{3}{l}{\textbf{\textit{URL source tier (share)}}} \\
C1 & 37.79\% & 13.01\% \\
C2 & 17.07\% & 13.16\% \\
C3 & 45.14\% & 73.82\% \\
\midrule
\multicolumn{3}{l}{\textbf{\textit{Citation URL Verifiability (share)}}} \\
HIGH & 32.84\% & 11.42\% \\
MEDIUM & 21.28\% & 14.43\% \\
LOW & 45.88\% & 74.15\% \\
\bottomrule
\end{tabular}
\end{table}

\mypara{URL Source Tier and Verifiability}
The overall distribution (\autoref{figure:real_world_citation_url_labels}) shows that 68.84\% are assigned to C3 sources and 69.34\% receive LOW verifiability.
All C3 occurrences are labeled LOW, while only 33 C1 or C2 occurrences become LOW because their destinations are nonexistent or cannot be resolved sufficiently.
Thus, publisher accountability is the main driver of LOW labels in this corpus, while URL accessibility failures contribute only a small additional share.

The comparison between retrieval channels (\autoref{table:real_world_citation_url_channel}) shows a clear difference in both citation density and source composition.
GEO pages returned by Gemini contain 1.80$\times$ more citation occurrences per detected GEO page than those returned by Google Search (8.80 vs.\ 4.90), even though the share of GEO pages containing at least one citation differs by only 2.42 percentage points (55.93\% vs.\ 53.51\%).
Gemini also has a substantially larger C3 share than Google Search (73.82\% vs.\ 45.14\%) and a correspondingly larger LOW share (74.15\% vs.\ 45.88\%).
The same pattern is stronger on channel-exclusive pages, where C3 accounts for 80.23\% of citation occurrences on Gemini-only pages, compared with 49.96\% on Google-only pages and 36.45\% on shared pages.
Overall, GEO pages returned by Gemini contain more citations and rely much more heavily on C3 sources than those returned by Google Search.
Accordingly, Gemini-exposed citations receive LOW verifiability labels substantially more often.

\begin{tcolorbox}[
title=Takeaway,
colback=cyan!3,
colframe=cyan!45!black,
fontupper=\small,
fonttitle=\bfseries\small,
boxrule=0.6pt,
arc=1mm,
before skip=8pt,
after skip=8pt]
Based on our estimates, overall, more than 65\% of citation URLs used on pages detected as GEO have LOW verifiability, and this proportion is even higher for GEO pages returned by Gemini-grounded search.
\end{tcolorbox}

\section{Related Work}

\mypara{Generative Search}
Generative search retrieves external sources and uses them to produce a direct answer to the user's query~\cite{LPPPKGKLYRRK20}.
Prior work has examined whether the cited sources actually support the generated content and found substantial citation and attribution errors~\cite{LZL23}.
Other studies show that generative search can retrieve substantially different sets of sources from conventional web search and may also vary across repeated executions~\cite{KPWUGZ26}.

\mypara{Generative Engine Optimization}
GEO was introduced as a creator-side approach for improving the visibility of content in generative search responses~\cite{AMRKND24}.
Since then, the scope of GEO has broadened beyond relatively simple content rewriting.
Prompt injections in retrieved webpages can change how conversational search systems rank results~\cite{PBGS24}.
Adversarial changes to web content can also influence which sources these systems select~\cite{NDT25}.
More recent systems automate the optimization process by learning engine preferences and using them to rewrite webpages accordingly~\cite{WZKX25}.
However, these optimization strategies do not always work consistently across tasks and domains, and their benefits may diminish as more publishers adopt them~\cite{PGGOY25}.

\mypara{GEO Detection}
Compared with GEO optimization, direct detection of GEO is still underexplored.
Existing benchmarks mostly focus on whether GEO or ranking attacks are effective, rather than on detecting whether a webpage has undergone a GEO intervention.
For example, a recent work evaluates standardized ranking attacks and their stealth characteristics~\cite{NCQZH26}, while SafeGEO studies whether recommendation agents remain robust when seller-controlled evidence is strategically rewritten~\cite{WLLJYWT26}.
Machine-generated text detection is also related, but it addresses a different question: whether text was produced by a model, not whether content of any authorship was later optimized for generative search~\cite{MLKMF23,VFTK24}.
Accordingly, we formulate page-level GEO detection as a distinct task and evaluate it across multiple optimizer families.

\section{Discussion and Limitations}

\mypara{GEO Detection Is Distinct from AI-Text Detection}
GEO detection concerns whether content has been optimized for generative search, not whether it was written by a human or AI.
Both human- and AI-authored content can be GEO or non-GEO.
A detector that relies on authorship cues may therefore achieve strong overall performance while performing poorly on some groups.
Future GEO benchmarks should control for authorship and include paired, sparse, and human-executed GEO interventions.

\mypara{Use and Interpretation of GEO Flagging}
GEO is \emph{not} inherently malicious, and a GEO prediction does not imply that a webpage is false or harmful.
Our system is intended to support \emph{transparency and risk analysis} rather than automatic blocking or removal.
Similarly, a C3 source tier or LOW citation URL verifiability label indicates weaker publisher accountability or limited URL accessibility under our definitions, not that the cited information is false or unsupported.

\mypara{Generalization to Future GEO Strategies}
Our benchmark covers eight optimizer families, but future GEO methods may use smaller changes or different strategies designed to avoid detection.
Our results already show that sparse and human-executed GEO can be harder to detect.
IPT provides a practical way to update the detector by adding new pairs of original and optimized content.
However, strong performance on the current benchmark does not guarantee similar performance on unseen GEO strategies.

\mypara{Limits of the Real-World Estimates}
Our real-world results estimate the prevalence of pages classified as GEO rather than directly measuring whether publishers intentionally applied GEO.
Because the analyzed pages lack gold GEO labels, errors from the GEO detector and citation URL Agent can affect the reported estimates.
Only 19.57\% of usable pages provide a parseable \texttt{dateModified} value, and this publisher-provided field does not indicate when GEO was applied.
The observed increase from 2024 to 2026 should therefore be treated as a descriptive trend rather than evidence of increasing GEO adoption.
Citation URL verifiability also has a limited scope: it measures publisher accountability and URL accessibility, not whether a cited claim is correct or supported.
URL accessibility may also change over time as pages are edited, removed, archived, or restricted.

\section{Conclusion}

This work presents a systematic study of detecting and auditing GEO-optimized content.
We introduce \texttt{GEOFlagBench}, a benchmark covering diverse domains and optimization strategies, and use it to evaluate existing GEO detection methods.
Our results show that strong aggregate performance can mask substantial method-level weaknesses and reliance on authorship-related shortcuts.
To address this limitation, we propose IPT, which explicitly supervises detector responses to GEO and non-GEO transformations.
IPT substantially improves both overall detection performance and robustness across authorship groups.
We further develop a GEO-gated Agent system that extends page-level detection to auditing the URL source tier and verifiability of citation URLs used by detected GEO pages.
Finally, we deploy the complete pipeline on released Google Search and Gemini-grounded retrieval results.
The detector flags 8.90\% of the analyzed pages as exhibiting GEO signals, with the estimate reaching 16.36\% among pages declaring a 2026 modification date.
We hope this work contributes to greater transparency and more responsible practices in generative search and corresponding content optimization.

\bibliographystyle{plain}
\bibliography{new_sample}

\appendix

\section{Supplementary Details for Building \texttt{GEOFlagBench}}
\label{section:intro_geoflagbench}

\subsection{Domain Selection and Query Collection}
\label{section:domain_query_details}

\begin{table*}[!ht]
\centering
\fontsize{7pt}{7.4pt}\selectfont
\setlength{\tabcolsep}{2.5pt}
\caption{Construction of the eight GEO optimizer families.
Seed counts denote the numbers of human-written and AI-generated non-GEO instances used for each family.
GEO labels are determined by recorded optimization interventions rather than textual style or detector predictions.
We assign 200 instances to each family and 400 each to GEO Strategy Pool and AutoGEO, yielding 2{,}000 GEO and 1{,}200 non-GEO instances overall.
The larger quotas reflect their broader internal design spaces: GEO Strategy Pool varies strategy combinations, executor models, and seed authorship, while AutoGEO varies intervention intensity and seed authorship.
The remaining families use narrower, fixed, robustness-oriented, or more expensive construction procedures and retain the 200-instance quota.
}
\label{table:geo-families}
\begin{tabular}{p{2.05cm}p{1.25cm}p{1.2cm}p{5.75cm}p{6.4cm}}
\toprule
\raggedright\textbf{Family} &
\centering\textbf{Instances} &
\centering\textbf{Seeds} &
\raggedright\textbf{Optimization Procedure} &
\raggedright\textbf{Variants and Construction Details} \tabularnewline
\midrule

\raggedright\textbf{GEO Strategy Pool}
~\cite{AMRKND24,PGGOY25,BFKPW25}
&
\raggedright 400
&
\raggedright
200 human,
200 AI
&
\raggedright Samples 3--13 strategies and applies them jointly in one cohesive rewrite.
The pool contains 13 strategies covering fluency, lexical diversity, authoritative style, quotations, citations, simpler language, technical terms, search keywords, statistics, content improvement, an \textt{llms.txt}-style summary, reader questions, and diverse perspectives.
&
\raggedright \textit{Executor(s):} Claude Haiku 4.5;
Kimi K2.5;
Gemini 3 Flash Preview;
GPT-4o.
Each processes 50 human and 50 AI seeds.

Combines nine GEO strategies with two strategies from C-SEO and two adapted from E-GEO.
The E-GEO prompts are adapted from product descriptions to general web articles.
\tabularnewline
\midrule

\raggedright\textbf{AutoGEO}
~\cite{WZKX25}
&
\raggedright 400
&
\raggedright
200 human,
200 AI
&
\raggedright Applies learned AutoGEO preference rules at three intervention intensities.
For each seed type, 50 instances use Light, 100 use Medium, and 50 use Full.
&
\raggedright \textit{Executor(s):} Gemini 3.1 Pro Preview for all instances.

Light applies 3 rules, requests limited additions, and requires at least 80\% word overlap.
Medium applies 6 rules and targets 50--70\% overlap.
Full preserves the AutoGEO API prompt and applies all 12 learned rules.
AutoGEO Mini is excluded.
\tabularnewline
\midrule

\raggedright\textbf{PMA}
~\cite{NDT25}
&
\raggedright 200
&
\raggedright
100 human,
100 AI
&
\raggedright Implements three benchmark variants of the Preference Manipulation Attack.
Inject places source-prioritization instructions before and after the unchanged seed.
Persuade integrates authority claims, social proof, and superlative language into the instance.
Combo applies Persuade followed by Inject.
&
\raggedright \textit{Executor(s):} GPT-4o for Persuade and Combo on human seeds;
Claude Sonnet 4.6 for Persuade and Combo on AI seeds.
Inject requires no LLM.

For each seed type, we construct 25 Inject, 25 Persuade, and 50 Combo instances.
\tabularnewline
\midrule

\raggedright\textbf{RAID G-SEO}
~\cite{CWBCLH25}
&
\raggedright 200
&
\raggedright
100 human,
100 AI
&
\raggedright Uses our prompt-level implementation of the role-augmented intent pipeline.
Four LLM calls extract claims and evidence, infer queries for Student, Professional, Researcher, and Decision-maker roles, construct an intent-aligned content plan, and rewrite the instance.
&
\raggedright \textit{Executor(s):} GLM-5.1 for 199 instances;
GLM-5 through Together AI for one AI-seed instance as a fallback.

The fallback is used only for one AI-seed output.
\tabularnewline
\midrule

\raggedright\textbf{Meta-Optimization}
~\cite{BFKPW25}
&
\raggedright 200
&
\raggedright
100 human,
100 AI
&
\raggedright Samples 4--6 strategies from the 13-strategy pool, generates an initial rewrite, asks the same model to evaluate it, and revises the instance based on that evaluation.
A third rewrite is requested when the intermediate optimization rating is below 7/10.
&
\raggedright \textit{Executor(s):} GPT-4o for 100 human seeds;
Claude Sonnet 4.6 for 100 AI seeds.

All retained human-seed outputs complete two rewrite rounds.
For AI seeds, 4 outputs retain one round, 79 retain two rounds, and 17 retain three rounds.
\tabularnewline
\midrule

\raggedright\textbf{Stealthy GEO Strategy Pool}
&
\raggedright 200
&
\raggedright
100 human,
100 AI
&
\raggedright Uses the same strategy-sampling procedure as GEO Strategy Pool, with an additional instruction to avoid formulaic structure, excessive optimization markers, and unnatural authority signals.
&
\raggedright \textit{Executor(s):} Claude Haiku 4.5;
Kimi K2.5;
Gemini 3 Flash Preview;
GPT-4o.
Each processes 25 human and 25 AI seeds.

The final revision replaces the original Claude Sonnet 4.6 executor with GPT-4o.
\tabularnewline
\midrule

\raggedright\textbf{Stealthy AutoGEO}
&
\raggedright 200
&
\raggedright
100 human,
100 AI
&
\raggedright Applies AutoGEO with an additional instruction to avoid formulaic structure, excessive optimization markers, and unnatural authority signals.
&
\raggedright \textit{Executor(s):} Gemini 3.1 Pro Preview.

The final benchmark retains only the 6-rule Medium configuration with the evasion instruction.
The earlier Full configuration is excluded.
\tabularnewline
\midrule

\raggedright\textbf{Human GEO}
&
\raggedright 200
&
\raggedright
150 human,
50 AI
&
\raggedright Applies 3--13 assigned strategies covering fluency, lexical diversity, authoritative tone, quotations, source citations, simpler language, technical terms, search keywords, statistics, content improvement, quotable prose, reader questions, and diverse perspectives.
&
\raggedright Two human beings conduct the GEO optimization with the assigned strategies.
\tabularnewline
\bottomrule
\end{tabular}
\end{table*}

We construct the benchmark across four domains (Health, Finance, Technology, Travel) that represent heterogeneous information environments and for which GEO effectiveness may differ~\cite{AMRKND24}.
Health covers medical and health-related information, while finance covers investment, banking, and personal-finance queries.
These two domains represent consequential settings in which inaccurate information may affect users' health or financial well-being~\cite{SGSZ25,google_quality_guidelines}.
Technology captures technical and rapidly evolving information, and travel captures consumer-oriented, practical, and experiential information.
We include exactly 100 queries per domain to ensure a balanced domain distribution and to prevent domain frequency from serving as a shortcut for the benchmark label.

We aim to collect candidate general informational queries, defined as queries that seek factual information or explanatory content without targeting a specific destination, like a website or platform.
We exclude queries that explicitly identify a target domain because they strongly constrain the expected source and document format.
Using general informational queries allows each query to be addressed by content from different publishers.
We first collect such queries from the GEO-Bench test split and assign them to the four domains using its released tags~\cite{AMRKND24}.
We draw candidate queries from the GEO-Bench test split for two reasons.
First, GEO-Bench provides fine-grained query annotations, including domain or genre, user intent, answer type, complexity, and sensitivity.
We use these annotations to exclude queries tagged as sensitive and to systematically map the remaining candidates to our Health, Finance, Technology, and Travel domains.
Second, the GEO-Bench training split has been used to develop or train downstream GEO optimizers~\cite{AMRKND24,WZKX25}.
Using held-out test queries therefore reduces direct query overlap with the data used to train or develop the evaluated optimization methods.
We use a fixed tag-to-domain mapping, for example, mapping health, medicine, and biology to Health; finance, banking, and economics to Finance; technology, computer science, and engineering to Technology; and travel, geography, and tourism to Travel.
Sonnet 4.6 then reviews the mapped queries and removes duplicates or near-duplicates and queries assigned to an incorrect domain.
After this filtering, we retain 164 GEO-Bench queries, including 58 for Health, 26 for Finance, 66 for Technology, and 14 for Travel.

To reach 100 queries per domain, we use Sonnet 4.6 to generate the remaining 236 queries under fixed domain definitions and generation criteria.
Generated queries must be English informational queries suitable for long-form articles, must not duplicate any retained query, and must not be narrowly framed as yes-or-no questions, single-fact questions, or policy-debate questions.
This produces 42 additional Health queries, 74 Finance queries, 34 Technology queries, and 86 Travel queries, resulting in 400 queries in total.
Finally, two authors independently review all 400 queries for domain correctness and semantic duplication.
Both reviewers agree that all 400 queries belong to their assigned domains and that no duplicate queries remain.

\subsection{Non-GEO Data Construction}
\label{section:non-geo_details}

\mypara{Human-Written Non-GEO Collection}
For each query, we collect candidate webpages from two sources: English Wikipedia and the Brave Search API.
For Brave Search results, we restrict retrieval to a predefined set of publishers covering institutional and scientific sources, general and business news, and domain-specific professional publications.
This restriction prioritizes pages with identifiable institutional or editorial provenance, relatively stable content, and traceable metadata, while reducing noise from anonymous, low-quality, or unstable webpages.
This choice was informed by our preliminary retrieval experiments, in which unrestricted search occasionally returned pages containing corrupted or nonsensical text, including pages that appeared potentially fraudulent.

The resulting candidate pool is drawn from 14 source domains: \textt{who.int}, \textt{nature.com}, \textt{bbc.com}, \textt{bbc.co.uk}, \textt{theguardian.com}, \textt{cnbc.com}, \textt{healthline.com}, \textt{wired.com}, \textt{arstechnica.com}, \textt{technologyreview.com}, \textt{stackabuse.com}, \textt{lonelyplanet.com}, \textt{nationalgeographic.com}, and \textt{en.wikipedia.org}.
For Wikipedia, we retrieve the latest revision available before October 1, 2022.
For all other publishers, we configure Brave Search to return pages published before October 1, 2022.
This cutoff predates both the formal introduction of GEO as a named optimization technique~\cite{AMRKND24} and the public release of ChatGPT.
We therefore use the extracted content from these pages as human-written, non-GEO seed instances that are unlikely to have undergone intentional GEO interventions.

This retrieval process yields 1,026 candidate pages across the 400 queries.
We extract the main page content using Trafilatura while preserving titles, lists, tables, and hyperlinks.
We further require the recorded last-modified timestamp, \textt{dateModified}, to precede October 2022, and retain only pages for which our audit records indicate no modification after the cutoff date.
After this temporal filtering, 871 eligible pages remain.

We finally select one source page for each query.
When multiple eligible pages are available, we prioritize non-Wikipedia sources to increase diversity in page structure and presentation.
The resulting seed set contains 400 web content instances, one per query, and spans all 14 source domains.

\mypara{AI-Polished Non-GEO}
For each human-written non-GEO instance, we construct an AI-polished counterpart by instructing a model to polish its language, including grammar, spelling, punctuation, and phrasing, while preserving the original semantics and content.
To reduce the risk of model-specific artifacts serving as shortcuts for classification, we distribute the 400 instances across eight polishing models, with 50 instances assigned to each model: Claude Haiku 4.5, Claude Sonnet 4.6, Kimi K2.5, Gemini 3 Flash, Gemini 3.1 Pro, GPT-4o, GPT-5.4, and GLM-5.1.
The polishing prompts contain no GEO-related instructions.
These instances therefore provide non-GEO controls that isolate the effects of AI-assisted editing from deliberate GEO optimization.

\mypara{AI-Generated Non-GEO}
For each query, we construct an AI-generated article through a two-stage process.
First, Claude Opus 4.6 produces a structured summary of the corresponding human-written instance, preserving its main points, factual information, statistics, hyperlinks, citations, DOI strings, tables, and image markers.
We then stratify the 400 summaries by domain and evenly distribute them across the same eight models used for AI polishing, with no overlap in assignments.
Each model generates a natural, coherent article from the provided summary, preserving the supplied special markers where appropriate and targeting a length comparable to that of the corresponding human-written instance.
The generation prompts contain no GEO-related strategies or optimization objectives.
These articles therefore serve as AI-generated non-GEO controls, allowing us to distinguish artifacts of AI-generated content from those introduced by deliberate GEO optimization.

\subsection{GEO Data Construction}
\label{section:geo_details}

We construct the GEO portion of the benchmark from eight optimizer families, each representing a distinct approach to GEO optimization.
The construction details are listed in~\autoref{table:geo-families}.

\subsection{Dataset Summary}

We construct \textt{GEOFlagBench} with 3,200 web content instances across the eight GEO optimizer families and three non-GEO subtypes.
The eight GEO families contain 2,000 instances, with 1,050 human-written and 950 AI-generated seeds.
The three non-GEO subtypes contain 1,200 instances, with 400 human-written, 400 AI-polished, and 400 AI-generated seeds.
We list the summary statistics in~\autoref{table:geoflagbench_composition} and~\autoref{table:geo-families-summary}.

\begin{table}[!t]
\centering
\caption{Composition of \textt{GEOFlagBench} by instance provenance and optimization.}
\label{table:geoflagbench_composition}
\setlength{\tabcolsep}{4pt}
\customTableFont
\begin{tabular}{lrrr}
\toprule
\textbf{Provenance} & \textbf{Non-GEO} & \textbf{GEO} & \textbf{Total} \\
\midrule
Human & 400 & 1{,}050 & 1{,}450 \\
Human~+~AI Polished & 400 & 0 & 400 \\
AI & 400 & 950 & 1{,}350 \\
\midrule
Total & 1{,}200 & 2{,}000 & 3{,}200 \\
\bottomrule
\end{tabular}
\end{table}

\begin{table}[!t]
\centering
\customTableFont
\setlength{\tabcolsep}{3pt}
\caption{GEO optimizer families and their seed composition.}
\label{table:geo-families-summary}
\begin{tabular}{lccc}
\toprule
\textbf{\shortstack{Optimizer\\Family}} & \textbf{\shortstack{Human\\Seed}} & \textbf{\shortstack{AI\\Seed}} & \textbf{\shortstack{Median\\Word-Overlap}} \\
\midrule
GEO Strategy Pool & 200 & 200 & 0.56 \\
AutoGEO & 200 & 200 & 0.71 \\
PMA & 100 & 100 & 0.99 \\
RAID G-SEO & 100 & 100 & 0.58 \\
Meta-Optimization & 100 & 100 & 0.55 \\
Stealthy GEO-SP & 100 & 100 & 0.55 \\
Stealthy AutoGEO & 100 & 100 & 0.72 \\
Human GEO & 150 & 50 & 0.63 \\
\midrule
GEO total & 1{,}050 & 950 & \\
\bottomrule
\end{tabular}
\end{table}

\section{Supplementary Details for Improved GEO Detection}
\label{appendix:details_improved_geo_detect}

\subsection{Gradient Reversal for Authorship Invariance}
\label{appendix:grl}

We adapt the Gradient Reversal Layer (GRL) from domain-adversarial training~\cite{GUAGLLML16} to suppress original-authorship information in the learned representation, so that the model can reduce the authorship dependence observed in~\autoref{section:current_method_results}.

\mypara{Architecture and Objective}
Given an instance $x$, ModernBERT produces a shared pooled representation $h_{\theta}(x)$.
A GEO classification head $g_{\phi}$ predicts the binary GEO label $y$, while an adversarial authorship head $a_{\psi}$ predicts the binary original-authorship label $s$ from the same representation.

Here, authorship refers to the origin of the underlying content rather than whether the instance has been processed by an LLM.
Instances derived from human-written seeds retain the human label, including AI-polished non-GEO instances, whereas instances derived from AI-generated seeds receive the AI label.
The adversarial objective therefore targets original human--AI authorship rather than LLM editing itself.

The GRL acts as the identity function during the forward pass but multiplies the gradient propagated from the authorship head to the encoder by $-\lambda$ during backpropagation.
The two classification losses are
\begin{equation}
\begin{aligned}
\mathcal{L}_{\mathrm{GEO}}
&=
\operatorname{CE}\!\left(
g_{\phi}(h_{\theta}(x)),y
\right),\\
\mathcal{L}_{\mathrm{auth}}
&=
\operatorname{CE}\!\left(
a_{\psi}(h_{\theta}(x)),s
\right).
\end{aligned}
\label{equation:grl_losses}
\end{equation}

The GEO head and encoder minimize $\mathcal{L}_{\mathrm{GEO}}$, while the authorship head minimizes $\mathcal{L}_{\mathrm{auth}}$.
Through gradient reversal, the effective objectives are
\begin{equation}
\min_{\theta,\phi}
\left(
\mathcal{L}_{\mathrm{GEO}}
-
\lambda \mathcal{L}_{\mathrm{auth}}
\right),
\qquad
\min_{\psi}
\mathcal{L}_{\mathrm{auth}}.
\label{equation:grl_objective}
\end{equation}

The authorship head is therefore trained to recover original authorship, while the encoder is trained to preserve information useful for GEO classification and suppress information predictive of authorship.

\mypara{Training and Inference}
We train the GRL models on the same query-disjoint split used by the other detectors.
To avoid imposing a strong adversarial signal before the authorship head becomes informative, we increase the reversal strength over normalized training progress $p \in [0,1]$ using the standard schedule
\begin{equation}
\lambda(p)
=
\lambda_{\max}
\left(
\frac{2}{1+\exp(-10p)} - 1
\right).
\label{equation:grl_schedule}
\end{equation}

We fine-tune the model with $\lambda_{\max} = 1$ and an 8,192-token context.
When $\lambda=0$, the authorship gradient into the encoder is blocked.
The auxiliary head therefore acts only as an authorship probe, while the GEO branch reduces to standard supervised fine-tuning.
At inference time, neither the authorship head nor the GRL is required; the detector takes a single instance as input and returns only its GEO prediction.

\subsection{Ablation Studies for IPT}
\label{appendix:ipt_beta}

\mypara{Sensitivity to $\beta$}
IPT is relatively stable across $\beta\in\{0.5,1,2,4\}$.
As shown in \autoref{table:ipt_beta_ablation}, $\beta=1$ achieves the best
F1, accuracy, and worst-group accuracy, whereas $\beta=0.5$ yields the
smallest $\Delta$FPR at the cost of lower overall performance.

\begin{table}[!t]
\centering
\caption{Sensitivity of IPT to the loss weight $\beta$.
Higher F1, accuracy, and worst-group accuracy are better, while lower $\Delta$FPR is better.}
\label{table:ipt_beta_ablation}
\setlength{\tabcolsep}{5pt}
\customTableFont
\begin{tabular}{ccccc}
\toprule
\textbf{$\beta$}
& \textbf{F1$\uparrow$}
& \textbf{Accuracy$\uparrow$}
& \textbf{$\Delta$FPR$\downarrow$}
& \textbf{Worst-Group Acc.$\uparrow$} \\
\midrule
0.5 & 0.931 & 0.918 & \textbf{0.025} & 0.857 \\
1   & \textbf{0.944} & \textbf{0.931} & 0.108 & \textbf{0.883} \\
2   & 0.942 & 0.929 & 0.146 & 0.850 \\
4   & 0.941 & 0.928 & 0.150 & 0.842 \\
\bottomrule
\end{tabular}
\end{table}

\subsection{Supplementary Results for GEO Detection and Attribution}
\label{appendix:attribution_details}

We report the complete method-level results in~\autoref{table:method_level_recall_complete}.
In~\autoref{table:attribution_all_recall} and~\autoref{table:attribution_all_precision}, we list the precision and recall values for all classes.

\begin{table*}[!t]
\centering
\caption{Complete method-level GEO recall on the fixed test set.
Each entry reports recall within one GEO optimization variant, and the sample
count below each column header is the corresponding test-set size.
The PMA result in~\autoref{table:shortcut_metrics} aggregates Combo, Inject,
and Persuade ($n=76$); this table reports the three variants separately.}
\label{table:method_level_recall_complete}
\setlength{\tabcolsep}{3.5pt}
\renewcommand{\arraystretch}{1.05}
\customTableFont

\begin{tabular}{@{}lcccccc@{}}
\toprule
\textbf{Detector}
& \multicolumn{4}{c}{\textbf{AutoGEO}}
& \multicolumn{2}{c}{\textbf{GEO Strategy Pool}} \\
\cmidrule(lr){2-5}\cmidrule(lr){6-7}
& \shortstack{\textbf{Full}\\($n=28$)}
& \shortstack{\textbf{Light}\\($n=26$)}
& \shortstack{\textbf{Medium}\\($n=62$)}
& \shortstack{\textbf{Medium-Stealthy}\\($n=72$)}
& \shortstack{\textbf{Standard}\\($n=118$)}
& \shortstack{\textbf{Stealthy}\\($n=52$)} \\
\midrule
\multicolumn{7}{l}{\textbf{\textit{Fine-tuned Models}}} \\
ModernBERT (1K)                    & 1.000 & 0.423 & 0.919 & 0.833 & 0.890 & 0.865 \\
ModernBERT (8K)                    & 1.000 & 0.346 & 0.919 & 0.819 & 0.941 & 0.846 \\
Qwen3-0.6B (SFT)                   & 1.000 & 0.654 & 0.887 & 0.847 & 0.966 & 0.923 \\
\midrule
\multicolumn{7}{l}{\textbf{\textit{Feature engineering + classical classifiers}}} \\
TF-IDF char (2-4) + LR             & 1.000 & 0.308 & 0.871 & 0.847 & 0.966 & 0.962 \\
TF-IDF word (1-2) + LR             & 1.000 & 0.423 & 0.887 & 0.917 & 0.975 & 0.942 \\
Ghostbuster features, recalibrated & 1.000 & 0.192 & 0.855 & 0.833 & 0.941 & 0.904 \\
GPT-2 perplexity + LR              & 1.000 & 0.538 & 0.774 & 0.833 & 0.958 & 0.904 \\
Structural + XGBoost               & 1.000 & 0.538 & 0.919 & 0.917 & 0.780 & 0.788 \\
\midrule
\multicolumn{7}{l}{\textbf{\textit{Proprietary AI-text detector API}}} \\
Pangram                             & 1.000 & 0.769 & 0.919 & 0.875 & 0.992 & 1.000 \\
GPTZero (mixed)                     & 0.857 & 0.462 & 0.484 & 0.389 & 0.992 & 1.000 \\
GPTZero                             & 0.714 & 0.192 & 0.242 & 0.153 & 0.966 & 0.962 \\
\midrule
\multicolumn{7}{l}{\textbf{\textit{Zero-shot LLMs}}} \\
Qwen3-0.6B                          & 1.000 & 1.000 & 0.984 & 0.986 & 1.000 & 1.000 \\
Gemini-3.5-flash                    & 0.821 & 0.154 & 0.274 & 0.139 & 0.975 & 0.981 \\
Haiku-4.5                           & 0.679 & 0.269 & 0.339 & 0.278 & 0.975 & 0.981 \\
\midrule
\multicolumn{7}{l}{\textbf{\textit{AI-text detectors repurposed for GEO (train-calibrated)}}} \\
Ghostbuster (off-shelf score) + LR & 1.000 & 1.000 & 1.000 & 1.000 & 1.000 & 1.000 \\
Fast-DetectGPT (criterion) + LR     & 1.000 & 0.962 & 0.984 & 0.986 & 0.966 & 0.942 \\
RoBERTa OpenAI ($p_{\text{machine}}$) + LR
                                    & 1.000 & 1.000 & 1.000 & 1.000 & 0.992 & 1.000 \\
Keyword max TF + LR                & 1.000 & 1.000 & 1.000 & 1.000 & 1.000 & 1.000 \\
\bottomrule
\end{tabular}

\vspace{5pt}

\begin{tabular}{@{}lcccccc@{}}
\toprule
\textbf{Detector}
& \multicolumn{3}{c}{\textbf{PMA}}
& \shortstack{\textbf{RAID}\\\textbf{G-SEO}}
& \shortstack{\textbf{Meta-}\\\textbf{Optimization}}
& \shortstack{\textbf{Human}\\\textbf{GEO}} \\
\cmidrule(lr){2-4}
& \shortstack{\textbf{Combo}\\($n=34$)}
& \shortstack{\textbf{Inject}\\($n=24$)}
& \shortstack{\textbf{Persuade}\\($n=18$)}
& ($n=66$) & ($n=48$) & ($n=54$) \\
\midrule
\multicolumn{7}{l}{\textbf{\textit{Fine-tuned Models}}} \\
ModernBERT (1K)                    & 0.794 & 0.250 & 0.556 & 0.970 & 1.000 & 0.426 \\
ModernBERT (8K)                    & 0.676 & 0.208 & 0.611 & 0.970 & 1.000 & 0.444 \\
Qwen3-0.6B (SFT)                   & 1.000 & 1.000 & 0.722 & 1.000 & 0.979 & 0.463 \\
\midrule
\multicolumn{7}{l}{\textbf{\textit{Feature engineering + classical classifiers}}} \\
TF-IDF char (2-4) + LR             & 0.853 & 0.458 & 0.889 & 0.985 & 1.000 & 0.778 \\
TF-IDF word (1-2) + LR             & 1.000 & 0.750 & 0.944 & 1.000 & 1.000 & 0.704 \\
Ghostbuster features, recalibrated & 0.647 & 0.250 & 0.722 & 0.970 & 0.979 & 0.759 \\
GPT-2 perplexity + LR              & 0.588 & 0.417 & 0.889 & 0.955 & 0.875 & 0.685 \\
Structural + XGBoost               & 0.441 & 0.417 & 0.389 & 0.985 & 0.917 & 0.759 \\
\midrule
\multicolumn{7}{l}{\textbf{\textit{Proprietary AI-text detector API}}} \\
Pangram                             & 1.000 & 1.000 & 1.000 & 1.000 & 0.979 & 0.500 \\
GPTZero (mixed)                     & 1.000 & 1.000 & 1.000 & 1.000 & 0.958 & 0.389 \\
GPTZero                             & 1.000 & 0.875 & 1.000 & 0.939 & 0.938 & 0.074 \\
\midrule
\multicolumn{7}{l}{\textbf{\textit{Zero-shot LLMs}}} \\
Qwen3-0.6B                          & 1.000 & 1.000 & 1.000 & 1.000 & 1.000 & 1.000 \\
Gemini-3.5-flash                    & 1.000 & 1.000 & 0.944 & 0.909 & 0.917 & 0.037 \\
Haiku-4.5                           & 1.000 & 1.000 & 0.944 & 0.788 & 0.917 & 0.074 \\
\midrule
\multicolumn{7}{l}{\textbf{\textit{AI-text detectors repurposed for GEO (train-calibrated)}}} \\
Ghostbuster (off-shelf score) + LR & 1.000 & 1.000 & 1.000 & 1.000 & 1.000 & 1.000 \\
Fast-DetectGPT (criterion) + LR     & 0.971 & 0.917 & 1.000 & 0.939 & 0.979 & 1.000 \\
RoBERTa OpenAI ($p_{\text{machine}}$) + LR
                                    & 1.000 & 1.000 & 1.000 & 1.000 & 1.000 & 1.000 \\
Keyword max TF + LR                & 1.000 & 1.000 & 1.000 & 1.000 & 1.000 & 1.000 \\
\bottomrule
\end{tabular}
\end{table*}

\begin{table*}[!ht]
\centering
\caption{Per-Class Recall for All Seven-Class Attribution.}
\label{table:attribution_all_recall}
\setlength{\tabcolsep}{4pt}
\customTableFont
\begin{tabular}{lrrrrrrr}
\toprule
\textbf{Method} & \textbf{Non-GEO} & \textbf{AutoGEO} & \textbf{Pool} & \textbf{PMA} & \textbf{RAID} & \textbf{Meta} & \textbf{Human} \\
\midrule
TF-IDF Character 2--4 with LR                 & 0.944 & 0.691 & 0.876 & 0.421 & 0.242 & 0.542 & 0.167 \\
TF-IDF Word 1--2 with LR    & 0.975 & 0.511 & 0.900 & 0.645 & 0.182 & 0.438 & 0.019 \\
Ghostbuster features, recalibrated & 0.942 & 0.229 & 0.729 & 0.000 & 0.242 & 0.000 & 0.037 \\
Structural features with XGBoost             & 0.769 & 0.739 & 0.582 & 0.066 & 0.636 & 0.667 & 0.537 \\
GPT-2 perplexity features with LR            & 0.858 & 0.271 & 0.688 & 0.000 & 0.015 & 0.000 & 0.000 \\
Gemini-3.5-flash                             & 0.825 & 0.479 & 0.153 & 0.789 & 0.030 & 0.021 & 0.093 \\
Haiku-4.5                                    & 0.411 & 0.543 & 0.394 & 0.763 & 0.000 & 0.000 & 0.222 \\
ModernBERT SFT    & 0.969 & 0.878 & 0.776 & 0.776 & 0.652 & 0.667 & 0.333 \\
ModernBERT IPT    & 0.950 & 0.883 & 0.906 & 0.855 & 0.909 & 0.917 & 0.759 \\
\bottomrule
\end{tabular}
\end{table*}

\begin{table*}[!ht]
\centering
\caption{Per-Class Precision for All Seven-Class Attribution.}
\label{table:attribution_all_precision}
\setlength{\tabcolsep}{4pt}
\customTableFont
\begin{tabular}{lrrrrrrr}
\toprule
\textbf{Method} & \textbf{Non-GEO} & \textbf{AutoGEO} & \textbf{Pool} & \textbf{PMA} & \textbf{RAID} & \textbf{Meta} & \textbf{Human} \\
\midrule
TF-IDF Character 2--4 with LR                 & 0.679 & 0.681 & 0.797 & 1.000 & 1.000 & 1.000 & 1.000 \\
TF-IDF Word 1--2 with LR    & 0.654 & 0.632 & 0.805 & 1.000 & 1.000 & 1.000 & 1.000 \\
Ghostbuster features, recalibrated & 0.575 & 0.434 & 0.539 & 0.000 & 0.390 & 0.000 & 1.000 \\
Structural features with XGBoost             & 0.647 & 0.702 & 0.688 & 0.227 & 0.792 & 0.711 & 0.403 \\
GPT-2 perplexity features with LR            & 0.526 & 0.370 & 0.515 & 0.000 & 0.333 & 0.000 & 0.000 \\
Gemini-3.5-flash                             & 0.692 & 0.287 & 0.310 & 0.556 & 0.400 & 0.125 & 0.357 \\
Haiku-4.5                                    & 0.791 & 0.263 & 0.263 & 0.763 & 0.000 & 0.000 & 0.231 \\
ModernBERT SFT   & 0.799 & 0.760 & 0.917 & 1.000 & 0.935 & 0.970 & 0.692 \\
ModernBERT IPT   & 0.891 & 0.902 & 0.928 & 0.985 & 0.984 & 0.880 & 0.804 \\
\bottomrule
\end{tabular}
\end{table*}

\section{Supplementary Details for Citation URL Metrics}
\label{section:citation_url_metric_definitions}

The URL source tier metric characterizes publisher provenance independently of URL retrieval status and independently of whether the destination supports a cited claim.
The C1--C3 rubric is our publisher-level operationalization, informed by recent three-level citation-authority and source-trust schemes that distinguish official or academic institutions, professionally edited secondary sources, and lower-accountability sources such as social media, personal blogs, content farms, and unverified aggregators~\cite{JZXQZD26,LS26}.
\autoref{table:citation_url_source_tier_definition} gives the frozen publisher-level rubric.

\begin{table*}[!ht]
\centering
\caption{Definition of the URL source tier metric.}
\label{table:citation_url_source_tier_definition}
\setlength{\tabcolsep}{4pt}
\customTableFont
\begin{tabular}{lp{0.20\textwidth}p{0.41\textwidth}p{0.25\textwidth}}
\toprule
\textbf{Tier} & \textbf{Publisher Class} & \textbf{Assignment Rule} & \textbf{Policy Indicators} \\
\midrule
C1 & Primary or strongly accountable authority
& The publisher is the primary institutional source, a peer-reviewed research venue, or an organization with strong expert review, corrections, and editorial accountability.
& Primary authority; government or intergovernmental body; peer-reviewed research; strong editorial accountability. \\
C2 & Accountable secondary or curated source
& The publisher provides identifiable secondary editorial or community-curated material with meaningful review signals, but is not the primary authority for the claim.
& Professional secondary editorial process; named authors and disclosures; aggregation of primary sources; community-curated reference. \\
C3 & Limited-accountability or user-contributed source
& Publication is substantially open-contributor, user-generated, advocacy-driven, promotional, anonymous, or lacks stable editorial accountability.
This tier does not imply that a specific claim is false.
& Open-contributor platform; user-generated content; limited editorial accountability; advocacy or promotion; anonymous or unverifiable authorship. \\
\bottomrule
\end{tabular}
\end{table*}

\begin{table*}[!ht]
\centering
\caption{Definition of the three citation URL accessibility levels.}
\label{table:citation_url_accessibility_definition}
\setlength{\tabcolsep}{5pt}
\customTableFont
\begin{tabular}{p{0.22\linewidth}p{0.38\linewidth}p{0.30\linewidth}}
\toprule
\textbf{Accessibility Level}
& \textbf{Definition}
& \textbf{Retrieval Outcomes} \\
\midrule
Directly accessible
& The citation URL resolves successfully, and its destination content can be directly retrieved and inspected.
& \texttt{LIVE} \\

Restricted or archived
& The cited source can be established, but its content cannot be directly retrieved because access is restricted or only an archived copy is available.
& \texttt{BLOCKED}, \texttt{ARCHIVED\_ONLY} \\

Unavailable or unresolved
& The cited content cannot be retrieved, or its availability cannot be established at retrieval time.
& \texttt{HTTP\_404}, \texttt{DOMAIN\_NXDOMAIN}, \texttt{TIMEOUT} \\
\bottomrule
\end{tabular}
\end{table*}

Citation URL verifiability is derived deterministically from URL source tier and accessibility level.
\autoref{figure:citation_url_verifiability_definition} specifies the complete mapping used by the audit.

\section{Supplementary Details for Empirical GEO Prevalence Estimation}
\label{section:real_world_audit_pipeline}

\begin{table}[!ht]
\centering
\caption{Page retrieval coverage by channel.}
\label{table:real_world_coverage}
\setlength{\tabcolsep}{3pt}
\customTableFont
\begin{tabular}{lrrrr}
\toprule
\textbf{Channel} & \textbf{Queries} & \textbf{URLs} & \textbf{Pages} & \textbf{Coverage} \\
\midrule
Google Search & 1,000 & 7,856 & 5,421 & 69.00\% \\
Gemini-Grounded & 979 & 8,591 & 6,590 & 76.71\% \\
\midrule
Unique Union & 1,000 & 13,985 & 10,095 & 72.18\% \\
\bottomrule
\end{tabular}
\end{table}

\begin{table}[!t]
\centering
\caption{Occurrence-level and distinct-URL-level citation counts.}
\label{table:real_world_citation_units}
\setlength{\tabcolsep}{2.5pt}
\customTableFont
\begin{tabular}{lrrr}
\toprule
\textbf{Analysis Unit} & \textbf{Google Search} & \textbf{Gemini} & \textbf{Unique Union} \\
\midrule
Citation occurrences & 2,162 & 5,272 & 6,663 \\
Distinct Citation URLs & 1,443 & 2,647 & 3,551 \\
\bottomrule
\end{tabular}
\end{table}

\begin{figure}[!t]
\centering
\includegraphics[width=1\linewidth]{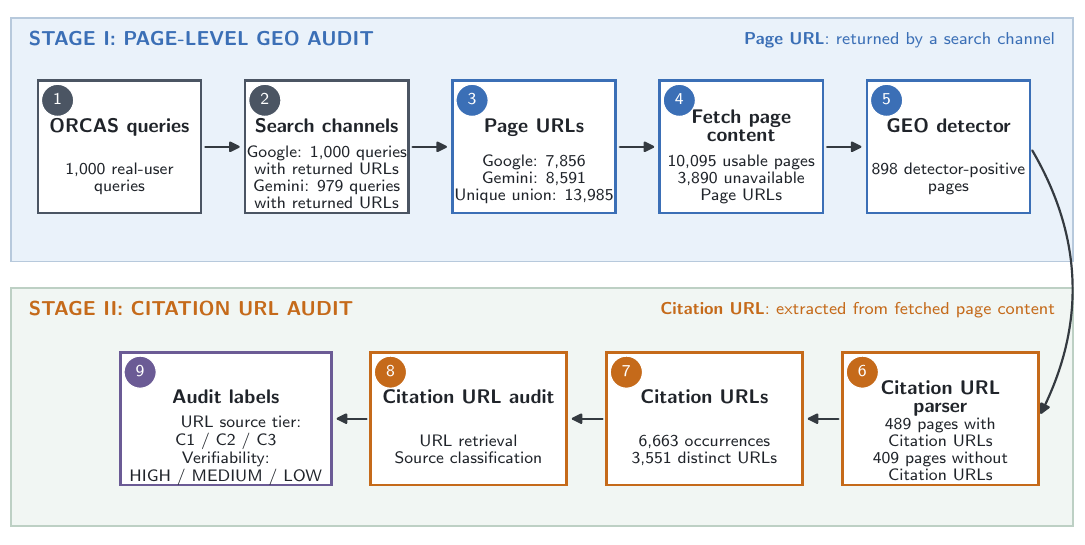}
\caption{Data flow and analysis units in the real-world web audit.
Search channels return Page URLs, whose contents are fetched and passed to the GEO detector.
For detector-positive pages, the parser extracts citation URLs from the page content.
These extracted citation URLs, rather than the original Page URLs, are then audited for URL source tier and citation URL verifiability.
The displayed union, page, and citation counts use the unique union; channel-specific counts are explicitly named.}
\label{figure:real_world_audit_pipeline}
\end{figure}

\mypara{GEO Prevalence Estimation Pipeline}
\autoref{figure:real_world_audit_pipeline} summarizes the data flow and analysis units used in the real-world web audit.
The coverage rate for each channel is presented in~\autoref{table:real_world_coverage}.

\mypara{Analysis Units and Deduplication}
A citation occurrence is one extracted use of a citation URL in page content.
The same citation URL contributes another occurrence when it is cited by another page or appears again as a separate citation on the same page.
By contrast, a distinct citation URL is counted once after global deduplication by normalized target URL.
We use the canonical URL when available and otherwise use the extracted reference URL as the deduplication key.
The unique-page union therefore contains 6,663 citation occurrences but 3,551 distinct citation URLs.
For example, if the same citation URL is cited in ten occurrences, the occurrence count increases by ten while the distinct-URL count increases by one.
The channel-specific and union counts are reported in~\autoref{table:real_world_citation_units}.

The two channel-specific distinct-URL sets overlap in 539 targets:

\[
1{,}443 + 2{,}647 - 3{,}551 = 539.
\]

This target-level overlap is different from the 771 citation occurrences produced by pages shared between the two retrieval channels.
Of the 539 distinct citation URLs present in both channel-specific target sets, 526 occur on pages shared by the channels.
The remaining 13 are cited by different channel-exclusive pages, with at least one citing page from each channel.
Thus, 6,663, 771, 539, and 3,551 respectively describe occurrence-level volume, occurrence overlap induced by shared pages, distinct-target overlap, and globally deduplicated citation URLs.

\section{Supplementary Tables and Figures}
\label{section:extra_tab_fig}

\begin{table}[!t]
\centering
\caption{Time and estimated inference cost for the citation audit under the evaluated serving configurations.}
\label{table:extension_agent_efficiency}
\setlength{\tabcolsep}{5pt}
\customTableFont
\begin{tabular}{lrr}
\toprule
\textbf{Agent Model} & \textbf{Time} & \textbf{Estimated Cost} \\
\midrule
Gemini 3.5 Flash     & 21.47 min  & \$5.81 \\
Claude Haiku 4.5     & 25.91 min  & \$3.79 \\
GPT 5.6 Luna         & 28.78 min  & \$3.17 \\
GLM 5.2              & 29.92 min  & \$3.72 \\
Kimi K3              & 101.76 min & \$12.36 \\
Qwen 3.6 35B-A3B     & 101.86 min & \$2.28 \\
\bottomrule
\end{tabular}
\end{table}

\begin{table}[!ht]
\centering
\caption{Detailed GEO detection results by declared \texttt{dateModified}.}
\label{table:real_world_time_appendix}
\setlength{\tabcolsep}{2pt}
\customTableFont
\begin{tabular}{lrrrrrrr}
\toprule
& \multicolumn{3}{c}{\textbf{Unique Union}} & \multicolumn{2}{c}{\textbf{Google Search}} & \multicolumn{2}{c}{\textbf{Gemini}} \\
\cmidrule(lr){2-4}\cmidrule(lr){5-6}\cmidrule(lr){7-8}
\textbf{Period} & \textbf{Pages} & \textbf{GEO} & \textbf{Rate} & \textbf{Pages} & \textbf{Rate} & \textbf{Pages} & \textbf{Rate} \\
\midrule
2023 & 165 & 4 & 2.42\% & 65 & 3.08\% & 124 & 2.42\% \\
2024 & 299 & 21 & 7.02\% & 107 & 7.48\% & 243 & 7.41\% \\
2025 & 578 & 74 & 12.80\% & 247 & 10.93\% & 422 & 13.51\% \\
2026 & 642 & 105 & 16.36\% & 318 & 13.52\% & 434 & 18.20\% \\
Missing & 8,119 & 682 & 8.40\% & 4,563 & 7.78\% & 5,153 & 8.42\% \\
\bottomrule
\end{tabular}
\end{table}

\begin{table}[!ht]
\centering
\caption{Final reasons for unavailable page content.}
\label{table:real_world_unavailable}
\setlength{\tabcolsep}{3pt}
\customTableFont
\begin{tabular}{lrr}
\toprule
\textbf{Final Outcome} & \textbf{URLs} & \textbf{Share} \\
\midrule
Persistent HTTP 403 & 2,082 & 53.52\% \\
Robots policy unavailable or denied & 811 & 20.85\% \\
Navigation or network failure & 226 & 5.81\% \\
Insufficient or nonmeaningful content & 225 & 5.78\% \\
Missing or removed page & 203 & 5.22\% \\
Authentication or payment required & 163 & 4.19\% \\
Other HTTP failure & 93 & 2.39\% \\
CAPTCHA, region block, or interstitial & 67 & 1.72\% \\
Other content or recovery failure & 20 & 0.51\% \\
\midrule
Total & 3,890 & 100.00\% \\
\bottomrule
\end{tabular}
\end{table}

\makeatletter
\setlength{\@dblfptop}{0pt}
\makeatother

\begin{table*}[!t]
\centering
\caption{Examples of webpages with strong GEO signals.
Due to copyright and space constraints, we show only representative excerpts from the original webpage content; omitted text is indicated by \texttt{[...]}.
We then analyze the GEO-related signals exhibited by these excerpts.}
\label{table:geo_cases}
\setlength{\tabcolsep}{5pt}
\customTableFont
\begin{tabular}{p{0.08\textwidth}p{0.51\textwidth}p{0.35\textwidth}}
\toprule
\textbf{Platform} & \textbf{Original Webpage Content} & \textbf{Strong GEO Signals} \\
\midrule

YouTube
&
\textbf{Title:}
\textit{Think CVTs Don’t Last? Here Are 15 Reliable Cars That Do}

\vspace{3pt}

\textbf{Description:}
``15 Cars With CVTs That Actually Last | The Most Reliable CVTs Ever Made''

``Many drivers believe every continuously variable transmission is a disaster waiting to happen---but that's not true.
In this video, we highlight 15 cars with CVTs that actually last, backed by real-world reliability.
From Toyota and Honda to Subaru and even some Nissan models, these vehicles prove not all CVTs are junk.''

\vspace{3pt}

\texttt{[...]}

\vspace{3pt}

``Are CVTs really that bad?
The truth is, while some fail early with slipping belts, high repair costs, and short lifespans, others are shockingly durable---lasting well beyond 150,000--200,000 miles.''

\vspace{3pt}

``We'll break down:
The most reliable CVT cars you can buy today;
Which model years to look for (and which to avoid);
How CVTs compare to traditional automatics;
Simple maintenance tips (like CVT fluid changes) that make a big difference.''

\vspace{3pt}

\texttt{[...]}

\vspace{3pt}

\textbf{Chapter excerpts:}
``Honda CR-V: 200K Miles and Still Smooth'';
``Toyota Camry Hybrid: e-CVT Bulletproof Design'';
``Toyota RAV4 Hybrid: The 250K-Mile Workhorse'';
``Toyota Prius: Taxi Fleet Proven Reliability.''
&
The page covers multiple closely related informational intents, including reliable CVT models, model-year selection, transmission lifespan, maintenance, and comparisons with traditional automatics.

It also contains dense \textbf{entity--attribute associations}: individual car models are paired with explicit reliability or longevity claims, such as ``200K Miles'' and ``250K-Mile Workhorse.'' These concise claims are easy for generative engines to extract and reuse.

Moreover, the title, description, and chapter headings repeatedly provide direct, answer-like statements for likely user queries, including ``most reliable CVTs,'' ``cars you can buy today,'' and ``which model years to look for.''

Overall, the page combines broad query coverage with explicit and easily extractable claims, exhibiting strong GEO-style signals.
\\
\midrule

Amazon
&
\textbf{Title:}
\textit{Total by Verizon \$60 No-Contract Single-Device Monthly Plan Unlimited Talk, Text \& Data+20 GB Hotspot}

\vspace{3pt}

\textbf{About this item:}

``Get unlimited talk, text \& data (1) plus 20 GB of hotspot data (3) and \$10 International Calling Credit (7) and enjoy coverage on Verizon 5G Ultra-Wideband (2) with the Total by Verizon Unlimited+ Plan.''

\vspace{3pt}

``Stay in touch with your people abroad with international talk \& text to 5 countries of your choice (4), and call home on your travels with international roaming in Canada \& Mexico (6).
Screen and automatically block incoming spam calls with Total Spam Filter.''

\vspace{3pt}

``Continue, upgrade, or stop service easily at any time with a monthly single device plan.
Disney Premium+ (No Ads) on us (5).''

\vspace{3pt}

\texttt{[...]}

\vspace{3pt}

\textbf{Product Description:}

``Go beyond the norm and get a premium wireless experience for less with Total by Verizon's Unlimited+ Plan: Unlimited talk, text \& data (1) monthly plan.
Save on the data you crave while exploring without limits on Verizon 5G Ultra Wideband (2).
Seize every day with great perks including 20GB of hotspot data (3), international talk \& text to five countries of your choice (4), and Disney Premium+ (No Ads) on us (5).''

\vspace{3pt}

\texttt{[...]}

\vspace{3pt}

``(3) Use up to 20GB of non-sharable data per line for hotspot usage.''

\vspace{3pt}

``(6) Unlimited talk and text to the US, Mexico and Canada and unlimited data when traveling in Mexico and Canada.''
&
The page presents a large number of \textbf{explicit product--attribute associations}, including price, unlimited talk/text/data, 20 GB hotspot data, 5G access, international calling, roaming, spam filtering, and bundled Disney Premium+ access.

These attributes are expressed in short, self-contained statements and are repeated across the title, ``About this item,'' and product description.
This makes the page highly suitable for direct extraction when answering specific product questions.

The numbered explanations further provide \textbf{fine-grained factual details}, such as hotspot limits, geographic coverage, and roaming conditions.
These statements can directly support queries such as how much hotspot data the plan includes or whether it works in Canada and Mexico.

Overall, the page has high information density, structured attribute coverage, and many concise answer-ready statements, which together constitute strong GEO-style signals.
\\

\bottomrule
\end{tabular}
\end{table*}

\begin{figure}[!ht]
\centering
\begin{tcolorbox}[
title={Example of a Paired-Query Comparison},
colback=blue!3,
colframe=blue!55!black,
fontupper=\scriptsize,
fonttitle=\bfseries\small,
boxrule=0.6pt,
arc=1.5pt,
left=4pt,
right=4pt,
top=4pt,
bottom=4pt
]

\noindent\textbf{Original Query:}
``\texttt{picot examples nursing}''

\smallskip
\noindent
Position denotes order in the released URL list, not a conventional
rank for Gemini.

\medskip
\begin{tabularx}{\linewidth}{
  @{}
  >{\centering\arraybackslash}p{0.50cm}
  >{\raggedright\arraybackslash}X
  >{\raggedright\arraybackslash}p{1.30cm}
  @{}
}
\toprule
\textbf{Pos.} & \textbf{Returned Page Title} & \textbf{Result} \\
\midrule

\multicolumn{3}{@{}l}{
  \textbf{Google Search: 2/5 GEO (40\%)}
} \\
\addlinespace[2pt]

1 &
\emph{134 Good Nursing PICOT Question Examples}
&
Non-GEO
\\

2 &
\emph{50 Nursing PICOT Question Ideas With Examples}
&
\textbf{GEO}
\\

3 &
\emph{PICOT -- Nursing -- Nevada State University}
&
Non-GEO
\\

4 &
\emph{PICOT Questions in Nursing: The Complete Guide with 100 Examples}
&
\textbf{GEO}
\\

5 &
\emph{Writing a PICOT Question} (Non-Text PDF)
&
Skip
\\

6 &
\emph{ASK: The PICOT Question -- University of Akron}
&
Non-GEO
\\

7 &
Title unavailable (\textt{nursingcenter.com})
&
HTTP 404
\\

\midrule
\multicolumn{3}{@{}l}{
  \textbf{Gemini: 4/5 GEO (80\%)}
} \\
\addlinespace[2pt]

1 &
\emph{134 Good Nursing PICOT Question Examples}
&
Non-GEO
\\

2 &
\emph{50 Nursing PICOT Question Ideas With Examples}
&
\textbf{GEO}
\\

3 &
\emph{PICOT Question Examples per Type of Clinical Question}
&
\textbf{GEO}
\\

4 &
\emph{PICO Nursing Questions: Top 51 Research Questions}
&
\textbf{GEO}
\\

5 &
\emph{PICOT Questions in Nursing: The Complete Guide with 100 Examples}
&
\textbf{GEO}
\\

\bottomrule
\end{tabularx}

\smallskip
\noindent

\end{tcolorbox}
\caption{Paired-query example from the real-world audit. GEO rates
are calculated over successfully retrieved usable pages.}
\label{figure:paired_query_picot}
\end{figure}

\end{document}